\documentclass[journal]{IEEEtran}
\usepackage{graphicx}
\usepackage{amsmath}
\usepackage{amssymb}
\usepackage[ruled]{algorithm2e}
\usepackage{bbm}
\usepackage{multirow}
\usepackage{booktabs}
\usepackage[hyphens]{url}
\usepackage{setspace}
\usepackage{color}

\begin{document}

\title{SLOGAN: Handwriting Style Synthesis for Arbitrary-Length and Out-of-Vocabulary Text}

\author{Canjie~Luo, 
        Yuanzhi~Zhu, 
        Lianwen~Jin*, 
        Zhe~Li, 
        and~Dezhi~Peng
\thanks{*Corresponding author.}
\thanks{Canjie Luo, Yuanzhi Zhu, Zhe Li, and Dezhi Peng are with the South China University of Technology, Guangzhou 510641, China (e-mail: canjie.luo@gmail.com; z.yuanzhi@foxmail.com; zheli0205@foxmail.com; eedzpeng@mail.scut.edu.cn).}

\thanks{Lianwen Jin is with the School of Electronic and Information Engineering, South China University of Technology, Guangzhou 510641, China, and also with the Peng Cheng Laboratory, Shenzhen 518055, China (e-mail: eelwjin@scut.edu.cn).}}

\markboth{IEEE TRANSACTIONS ON NEURAL NETWORKS AND LEARNING SYSTEMS}
{Shell \MakeLowercase{\textit{et al.}}: Bare Demo of IEEEtran.cls for IEEE Journals}

\maketitle

\begin{abstract}
Large amounts of labeled data are urgently required for the training of robust text recognizers. However, collecting handwriting data of diverse styles, along with an immense lexicon, is considerably expensive. Although data synthesis is a promising way to relieve data hunger, two key issues of handwriting synthesis, namely, style representation and content embedding, remain unsolved. To this end, we propose a novel method that can synthesize parameterized and controllable handwriting {S}tyles for arbitrary-{L}ength and {O}ut-of-vocabulary text based on a {G}enerative {A}dversarial {N}etwork ({GAN}), termed {SLOGAN}. Specifically, we propose a style bank to parameterize the specific handwriting styles as latent vectors, which are input to a generator as style priors to achieve the corresponding handwritten styles. The training of the style bank requires only the writer identification of the source images, rather than attribute annotations. Moreover, we embed the text content by providing an easily obtainable printed style image, so that the diversity of the content can be flexibly achieved by changing the input printed image. Finally, the generator is guided by dual discriminators to handle both the handwriting characteristics that appear as separated characters and in a series of cursive joins. Our method can synthesize words that are not included in the training vocabulary and with various new styles. Extensive experiments have shown that high-quality text images with great style diversity and rich vocabulary can be synthesized using our method, thereby enhancing the robustness of the recognizer.
\end{abstract}


\IEEEpeerreviewmaketitle

\section{Introduction}

\IEEEPARstart{T}{he} written word, a remarkable human achievement, marks the transition from the days of prehistory to the contemporary written history~\cite{Fogel2020ScrabbleGAN}. Today, handwriting is recognized as a unique and essential capability of human beings~\cite{Choudhury2019Synthesis}. Even in today's digital era, handwritten text still has wide applications, including impromptu note-taking, mathematical operations, business transactions, and postal mail labeling. Owing to the ubiquity of handwritten words, offline handwritten text recognition is an important area in the field of computer vision.

\begin{figure}[t]
\centering
\includegraphics[width=1\columnwidth]{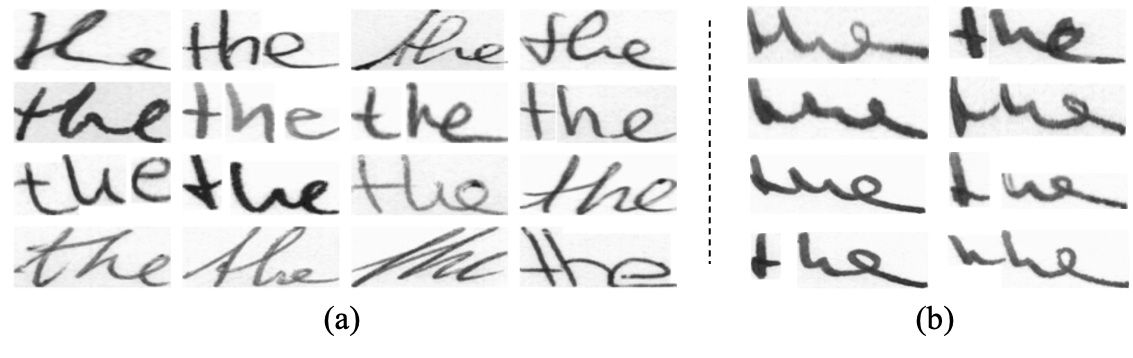} 
\caption{Samples collected from a widely used IAM dataset~\cite{marti2002iam}. Recognizing text in distinct individual handwriting styles is challenging, because (a) the handwriting styles of a word (for instance, ``the") written by different people can be significantly different, and (b) the style of the word changes observably every time the same person writes.}
\label{pic-word-the}
\end{figure}

\begin{figure}[t]
\centering
\includegraphics[width=0.85\columnwidth]{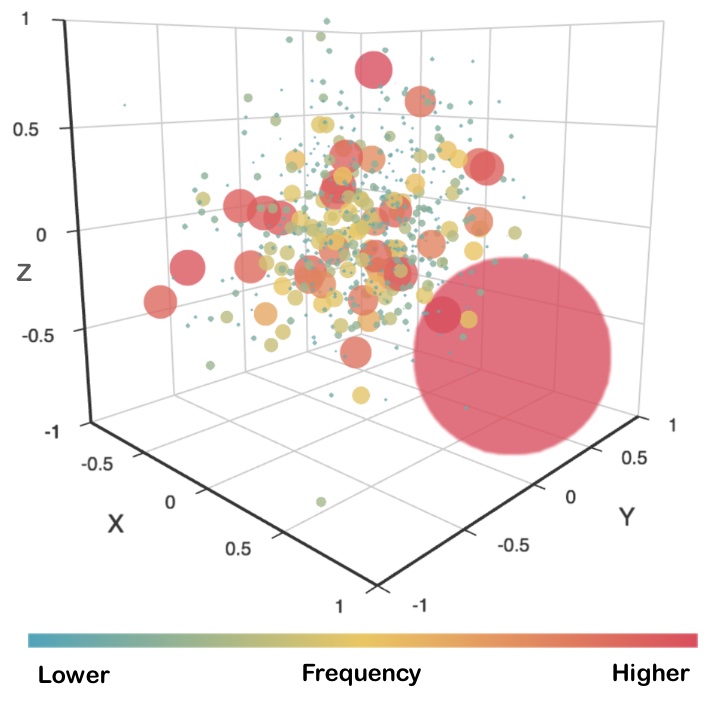} 
\caption{Imbalanced distribution of handwriting styles revealed using our method. The handwriting styles are from the popular IAM training set~\cite{marti2002iam}. Our method can parameterize the styles as vectors. The dimensionality of the style vector is set to three, so that each style can be visualized in a three-dimensional coordinate system. The color (and also the size) of every point denotes the frequency at which a specific style occurs in the dataset. There are significant biases in terms of both the style and frequency distribution. A vast expanse of empty space indicates a lack of styles.}
\label{pic-style-bias}
\end{figure}

Recent data-driven approaches~\cite{Xiao2019Pixel,Kang2020Unsupervised} have significantly improved the recognition performance. However, recognizing and processing of a plethora of distinct individual handwriting styles still remain a great challenge~\cite{Bhunia2019Handwriting,luo2020learn,liu2020stroke}. As shown in Figure~\ref{pic-word-the} (a), the handwriting styles of a word (\textit{e.g.}, ``the") written by different people can be significantly different. In fact, the style in which the word is written, changes observably every time the same person writes it, as shown in Figure~\ref{pic-word-the} (b). This suggests that the number of handwriting styles is almost limitless. Meanwhile, existing training data are insufficient and cannot represent all possible writing styles. To further illustrate the handwriting style distribution, we provide the styles in the popular IAM training set~\cite{marti2002iam} parameterized via our proposed method, in which the style vector dimensionality is set to three for visualization. As shown in Figure~\ref{pic-style-bias}, a wide empty space indicates a lack of handwriting styles. Simultaneously, we observed significant biases in terms of both the style and frequency distributions. For instance, the dataset contains 3,811 samples from Writer \#0, but only 10 samples from Writer \#6. 

One promising strategy for handling the absence of style is geometry augmentation, including feature-level augmentation (\textit{e.g.}~\cite{Bhunia2019Handwriting}) and image-level augmentation (\textit{e.g.}~\cite{luo2020learn}). However, augmentation is performed on the basis of existing samples, which means that Out-Of-Vocabulary (OOV) samples cannot be created. Another solution is to collect and annotate more images specifically for training. However, this method is time-consuming and labor-intensive.

Automatically generating handwritten text images can reverse the annotation process to save costs, by starting from a given word and generating a corresponding image of the handwritten text~\cite{Alonso2019Adversarial}. Previous studies~\cite{Graves2013Generating,Zhang2018Drawing,Kotani2020Decoupled} have made use of online handwriting data for training. A well-trained network can generate complex sequences with long-range structures by recurrently predicting data points in a step-by-step manner. Recently, non-recurrent generative methods~\cite{Fogel2020ScrabbleGAN,Alonso2019Adversarial,Kang2020GANwriting,gan2021higan} have been shown to directly produce a synthetic handwriting image according to a given text string. They also demonstrated the superiority of imitating offline handwriting features (vivid strokes, textures, paper background, \textit{etc.}).

\begin{figure}[t]
\centering
\includegraphics[width=0.9\columnwidth]{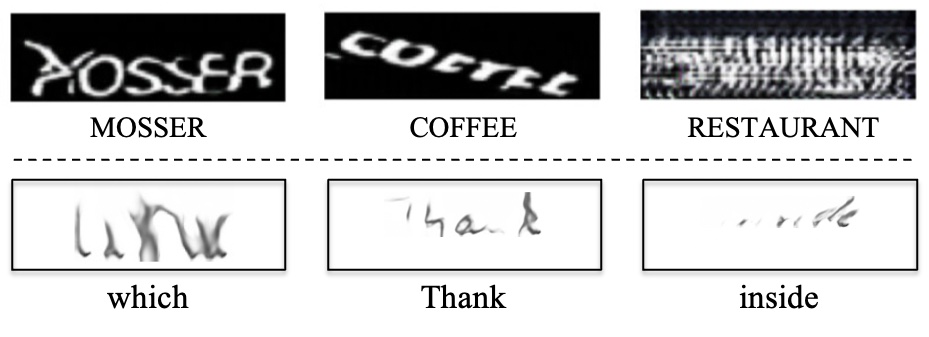} 
\caption{Failure cases of text images generated using vanilla domain-transfer GANs. Top: as reported by Luo \textit{et al.}~\cite{luo2021separating}, CycleGAN~\cite{Zhu2017cycleGAN} cannot preserve the strokes of every character in the task of removing noisy background from scene text images; bottom: Kang \textit{et al.}~\cite{Kang2020GANwriting} observed that even the advanced domain-transfer model ``FUNIT"~\cite{liu2019few} cannot generate relatively usable training samples. Unclear characters and unexpected artifacts indicate a lack of supervision and the confusion of the generator.}
\label{pic-vanilla-gan}
\end{figure}

Despite such impressive efforts, handwriting data synthesis remains challenging and unsolved. For instance, we found that the random noisy latent vector $\pmb{z} \sim \mathcal{N}(0, 1)$, which has been extensively employed in existing Generative Adversarial Networks (GANs)~\cite{brock2018large,karras2019style,Choi2020StarGAN}, is insufficient for modeling specific handwriting styles. Moreover, previous studies embedded content via complex modules, including recurrent embeddings~\cite{Alonso2019Adversarial}, character-wise and string-level encoders~\cite{Kang2020GANwriting}, a series of convolutional filters~\cite{Fogel2020ScrabbleGAN} and filter maps~\cite{gan2021higan}, which limit the flexibility and practicability of content input. For instance, the adjacent character interval cannot be easily adjusted~\cite{Fogel2020ScrabbleGAN,Alonso2019Adversarial,gan2021higan}. It is also impossible to generate a text string longer than the maximum number in the training phase~\cite{Alonso2019Adversarial,Kang2020GANwriting}.

To this end, we propose addressing the above-mentioned two key issues, the \textit{style representation} of the handwriting and the \textit{content embedding} of the text, as follows. 1) We propose a style bank to store parameterized handwriting style vectors, which are taken by the generator to guide the generated images toward specific styles. Our model is differentiable, thus, the style vectors can be updated by the joint training with the generator. Subsequently, new styles can be synthesized by controlling the latent style parameters. 2) Our method simplifies the content embedding design by providing only a printed style image as input, which is cheap to obtain and can be rendered online during training. After the training, it is possible to generate handwritten text images with diverse contents, including various adjacent character intervals, curved text, and arbitrary length text, by changing only the printed characters and their position on the input images.

\begin{figure}[t]
\centering
\includegraphics[width=1.\columnwidth]{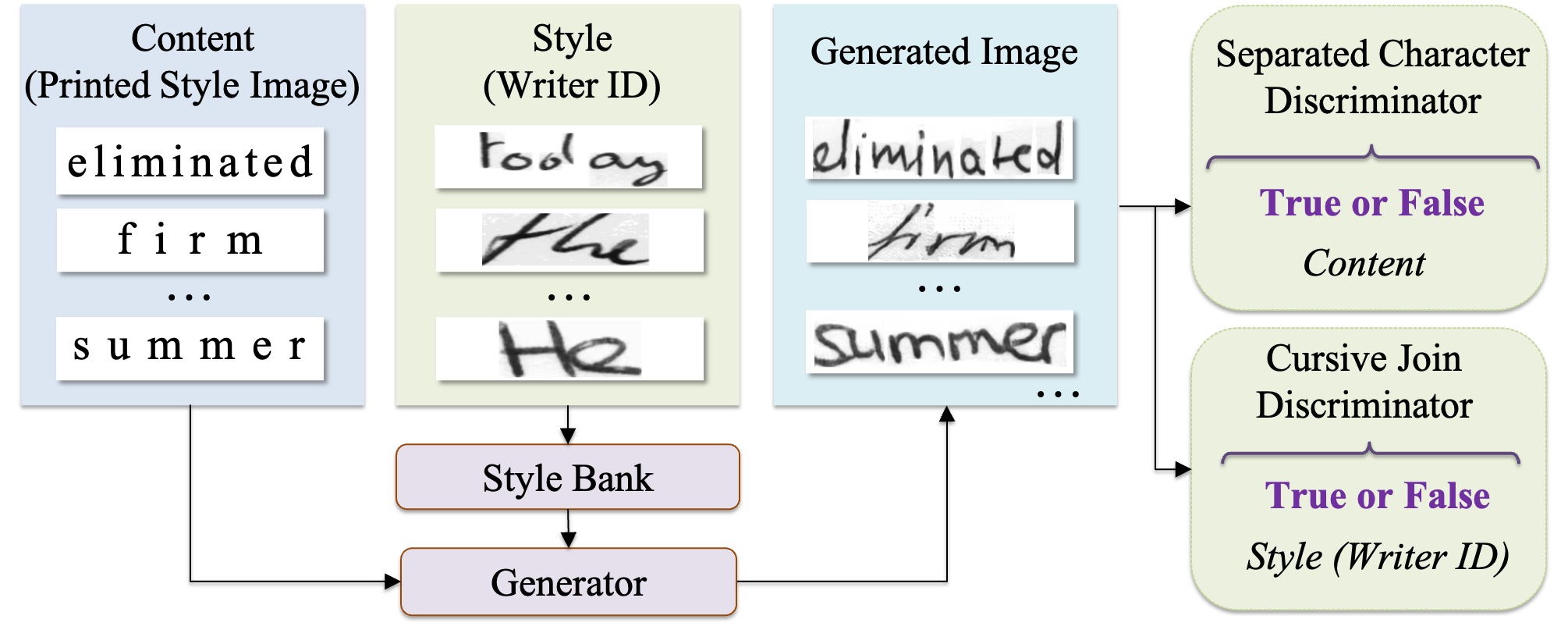} 
\caption{Learning scheme of the proposed method to imitate specific handwriting styles. The style bank embeds the writer ID into a latent vector, which is taken by the generator to transfer the printed style to the corresponding handwriting style. The style bank and the generator are jointly optimized under the supervision of the dual discriminators.}
\label{pic-learning-scheme}
\end{figure}

Although there are plenty advantages of modeling handwriting synthesis as a domain-transfer task (from printed style to handwritten style), existing GAN-based methods are not sufficiently robust to generate a text string image~\cite{Kang2020GANwriting,luo2021separating,Fang2019draw}, because one text string image contains multiple objects in a sequence~\cite{luo2019moran,zhang2019sequence}. For instance, it has been reported in previous studies~\cite{Kang2020GANwriting,luo2021separating} that the generated images contain unclear characters and unexpected artifacts, which indicates the lack of supervision and the confusion of the generator, as shown in Figure~\ref{pic-vanilla-gan}. As our goal is to propose an approach that writes like humans, we should review the nature that characterizes the human perception and learning systems. The cognitive science literature revealed that two reciprocal mechanisms, namely bottom-up and top-down processing,  complementarily form rich visual representations and create vivid imageries~\cite{gibson2002theory,intaite2013interaction}. The former focuses on local elements (parts of the whole) while the latter begins at a more global view, such as surrounding context and prior knowledge. Inspired by this finding, we design dual discriminators to guide the generator at two levels, namely the separated characters and the cursive joins between them. The learning scheme of the proposed method is illustrated in Figure~\ref{pic-learning-scheme}. By this way, our method can synthesize previously parameterized and now controllable handwriting {S}tyles for arbitrary-{L}ength and {O}ut-of-vocabulary text based on a {GAN} (SLOGAN).

To summarize, the contributions of this study are as follows:

\begin{itemize}
\item We propose parameterizing the handwriting styles as latent vectors, which serve as style priors to guide the generator to achieve specific handwriting styles. By manipulating the elements of the latent vector, a new handwriting style that does not exist in the dataset can be created. The training of the style parameterization requires only the writer IDs of the source images.
\item We propose embedding text content using a printed style image. Once trained, the proposed method can generate out-of-vocabulary words and sentences of arbitrary length by changing only the text string on the printed style text image. Furthermore, spatial diversity can also be achieved by introducing the variance in the arrangement of the character position. 
\item We propose dual discriminators to supervise the generator at the separated character level and the cursive join level, respectively, which can effectively improve the quality of the generated images. 
\item Extensive experiments show that the proposed method can generate verisimilar handwritten text images, which further enriches the diversity of existing training samples and improves the robustness and performance improvement of the recognizers.
\end{itemize}

\section{Related Work}
\subsection{Handwritten Text Image Synthesis} 

The past decade has witnessed significant advances in deep learning. Recurrent Neural Networks (RNNs)~\cite{hochreiter1997long,cho2014learning} have been widely used for modeling complex sequences with long-range structures~\cite{Graves2013Generating,Zhang2018Drawing,Kotani2020Decoupled}. The approaches made use of online handwriting data for training and predicted data points recurrently in a step-by-step manner. Nevertheless, the online sequence cannot represent certain offline handwriting features, such as stroke width, textures, and paper backgrounds. 

Image and texture synthesis are challenging tasks~\cite{zhang2018face,liu2019collocating}. With the breakthrough of GANs~\cite{Zhu2017cycleGAN,goodfellow2014generative,Mirza2014ConditionalGAN,zhao2020masked,you2020bayesian,Abdal2020Image2StyleGAN,yeo2021simple}, directly generating a handwritten text image has become an interesting topic. Non-recurrent generative methods~\cite{Fogel2020ScrabbleGAN,Alonso2019Adversarial,Kang2020GANwriting,gan2021higan}  can produce a handwritten text image according to a given text string. The generated images through such approaches are highly lifelike, with vivid strokes, textures and paper backgrounds. However, the two key issues of handwriting synthesis, namely, style representation and content embedding, are still challenging. 1) Alonso \textit{et al.}~\cite{Alonso2019Adversarial} and Fogel \textit{et al.}~\cite{Fogel2020ScrabbleGAN} simply followed the setting of popular GANs using a random noise latent vector. However, this approach may not sufficiently represent the variance of handwriting styles, which further limits style diversity. We argue that the latent vector can be more specific and effective for representing styles. This view is strongly supported by several facts. For instance, Lai \textit{et al.}~\cite{lai2020synsig2vec} extracted the underlying neuromuscular parameters of genuine signatures to eliminate the need for skilled forgeries. Moreover, with respect to natural scene objects, the generated images can be edited by enabling the interpretation of the GANs' latent space~\cite{Abdal2020Image2StyleGAN,voynov2020unsupervised,harkonen2020GANSpace}. This inspires us to find a way to parameterize the handwriting styles and introduce more controllable parameters to generate various styles. 2) Previous studies proposed complex modules to embed content, such as recurrent embeddings~\cite{Alonso2019Adversarial}, character-wise and string-level encoders~\cite{Kang2020GANwriting}, a series of convolutional filters~\cite{Fogel2020ScrabbleGAN} and filter maps~\cite{gan2021higan}. However, these modules have limited application because they are not sufficiently flexible to embed various contents. For instance, it is impossible to embed a text string longer than the maximum number in the training phase~\cite{Kang2020GANwriting}. The series of convolutional filters~\cite{Fogel2020ScrabbleGAN} and filter maps~\cite{gan2021higan} cannot flexibly adjust the adjacent character interval to achieve various character arrangements. Therefore, we provide the text content on a printed style image and feed it into the generator. By this way, we can generate various contents by changing the string and rearranging the characters on the input image.

\subsection{Handwritten Text Image Augmentation} 

Another promising way to improve the diversity of the training data is text image augmentation, which is typically achieved through the geometry augmentation. Wigington \textit{et al.}~\cite{Wigington2017Augmentation} augmented the existing text images using random perturbations on a regular grid. Bhunia \textit{et al.}~\cite{Bhunia2019Handwriting} elastically warped the extracted features in a scalable manner. Luo \textit{et al.}~\cite{luo2020learn} proposed a learnable augmentation method to obtain more effective and specific samples. These methods have made significant progress and considerably boost the performance of the recognizers.

However, images that are nonexistent in the training data, such as images containing out-of-vocabulary words, cannot be created through data augmentation. Thus, the augmentation is bounded by the existing dataset. The proposed data synthesis method can be the complement of data augmentation and further benefit the recognition robustness.

\subsection{Font Style Transfer} 

The font style (text effect) transfer is a subtopic of general image style transfer~\cite{johnson2016perceptual}. Recent works~\cite{Jiang2019SCFont,Li2019FETGAN,Yang2019Controllable,Zhu2020FewShot,Yang2020TE141K,Wang2020Attribute2Font,xi2020jointfontgan} have intensively studied the text effect on a printed single character and achieved great success. However, these approaches cannot be directly deployed for handwriting synthesis. First, almost all the font style transfer approaches require laborious annotations for supervision, such as paired training samples, which consist of input images and corresponding pixel-level aligned ground-truth images. Meanwhile, additional annotation of attributes is required to achieve editable attributes~\cite{Wang2020Attribute2Font}. Second, the approaches were designed for a few specific font effects. The number of font styles is relatively limited. For instance, the representative benchmark, TE141K~\cite{Yang2020TE141K}, comprised only 152 font effects, whereas the handwriting dataset IAM~\cite{marti2002iam} was collected from 657 writers. Third, previous studies focused on the effects of only a single character, however, a handwritten text image typically contains multiple characters. Finally, these approaches were proposed for design industry, rather than data synthesis for text recognition.

Unlike the above-mentioned font-transfer approaches, the proposed method can generate an image containing a long text string, instead of only a single character. Our motivation is to enrich the data diversity and improve the robustness of the text recognizers. In this regard, previous studies have only demonstrated the visual effects.

\section{Methodology}

\begin{figure*}[t]
\centering
\includegraphics[width=1.9\columnwidth]{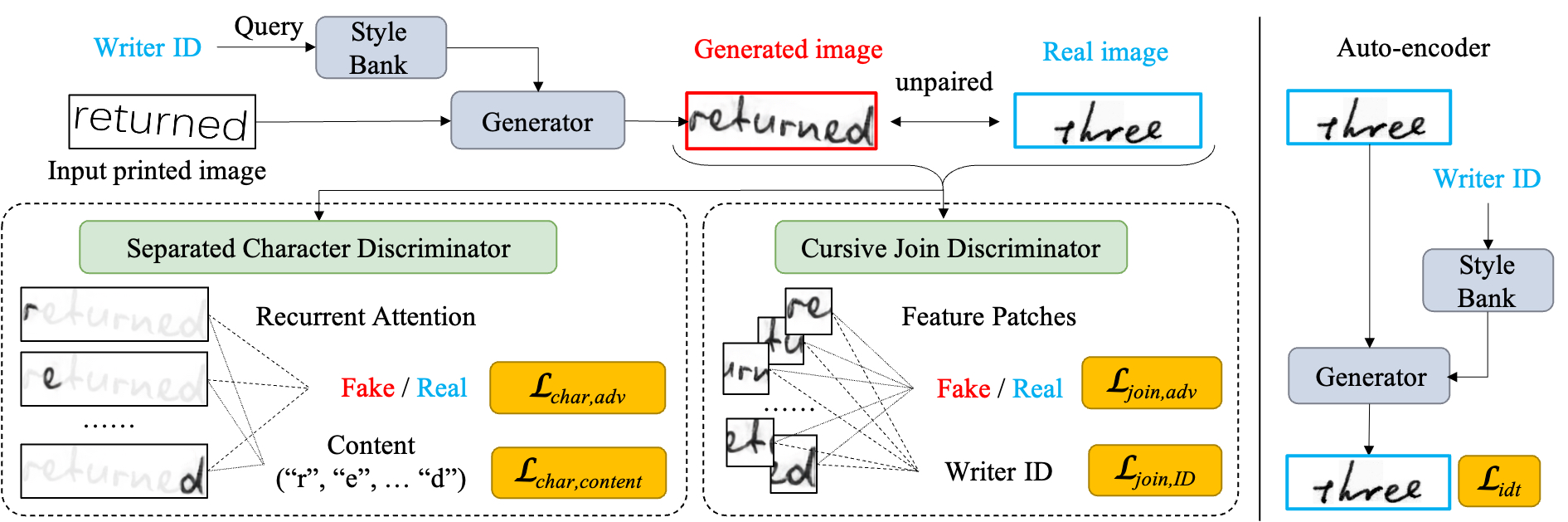} 
\caption{Our overall framework in the training phase. The style bank stores latent vectors representing handwriting styles. It is updated jointly with the generator. The generator is supervised by the dual discriminators, which guide the style transfer at character level and cursive join level, respectively. The generator and discriminators are trained in alternating steps.}
\label{pic-overall}
\end{figure*}

In this section, we first describe the proposed approach whereby handwriting styles are parameterized by imitating specific styles. This makes it possible to control parameters to generate various handwriting samples. Then, we present the design of dual discriminators, which are crucial to the success of our approach. Finally, we illustrate the inference of the generator for diverse styles and contents.

We provide an annotation table as below to facilitate reading and reference.

\begin{table}[h]
\centering
\footnotesize
\caption{Abbreviations and symbols.}
\label{table:annotation}
\setlength{\tabcolsep}{1.5mm}{
\begin{tabular}{ll}
\toprule
$I$, $Y$ & Image and its corresponding label  \\
\midrule
$G$ & Generator \\
$\operatorname{StyleBank}$ & Style bank \\
\midrule
$D_{\operatorname{char}}$ & Separated character discriminator  \\
$D_{\operatorname{char, adv}}$ & Adversarial training head of $D_{\operatorname{char}}$ \\
$D_{\operatorname{char, content}}$ & Content supervision head of $D_{\operatorname{char}}$ \\
\midrule
$D_{\operatorname{join}}$ & Cursive join discriminator  \\
$D_{\operatorname{join, adv}}$ & Adversarial training head of $D_{\operatorname{join}}$ \\
$D_{\operatorname{join, ID}}$ & Handwriting style supervision head of $D_{\operatorname{join}}$ \\
\bottomrule
\end{tabular}
}
\end{table}

\subsection{Handwriting Style Parameterization by Imitation}

As shown in Figure~\ref{pic-overall}, our framework consists of a style bank, a generator, and two discriminators. The generator takes the prior provided by the style bank and transfers the printed style to the corresponding handwriting style. Specifically, the style bank is a simple lookup table that stores $n$ handwriting styles as latent vectors $\pmb{z}_{all} \in \mathbb{R}^{d \times n}$, where $d$ is the dimensionality of a latent vector. Given a writer ID, the style bank returns the corresponding latent vector
\begin{equation}
\pmb{z} = \operatorname{StyleBank}(\operatorname{ID}),
\end{equation}
where the latent vector $\pmb{z}$ serves as the prior that guides the generator to achieve the target handwriting style. The style bank is randomly initialized and jointly updated with the generator under the supervision of the writer IDs. The generator $G$ is an encoder-decoder architecture, which takes the printed style image $I_{\operatorname{print}}$ as input and generates an image
\begin{equation}
{I_{\operatorname{fake}}} = G({I_{\operatorname{print}}}, \pmb{z}).
\end{equation}
Here, $I_{\operatorname{fake}}$ contains the same content as $I_{\operatorname{print}}$ and corresponds to the handwriting style of $\pmb{z}$.
Following the methods for scene text data synthesis~\cite{jaderberg2016reading,gupta2016synthetic}, we render printed style text on a white background to obtain $I_{\operatorname{print}}$, which can be easily rendered online during training.

After obtaining the generated image $I_{\operatorname{fake}}$, we design two discriminators to perform supervision at the separated character level and the cursive join level, respectively. This design is critical to the success of the image-to-image transfer of handwritten text image. The two discriminators are introduced as follows.

\textsl{1) Separated Character Discriminator (Adversarial Training and Character Content Supervision):} Although previous studies~\cite{Fogel2020ScrabbleGAN,Alonso2019Adversarial} presented promising results, the generated images contain unclear characters and unexpected artifacts, which indicates the generator still suffers under-fitting. We found that the image-level discrimination is inefficient for training. Thus, we design a discriminator $D_{\operatorname{char}}$ to supervise the generator at the character level. To address the lack of character-level bounding box annotation, we leverage the \textsl{attention mechanism}~\cite{bahdanau2014neural}, which is widely used for sequence-to-sequence mapping. It takes only a text string label as weak supervision to localize characters. After obtaining the character positions, we can further perform adversarial training and content (character category) supervision for every character. Specifically, the $D_{\operatorname{char}}$ consists of two heads, termed $D_{\operatorname{char, adv}}$ and $D_{\operatorname{char, content}}$. The $D_{\operatorname{char, content}}$ head leverages the \textsl{attention mechanism} to localize characters. The $D_{\operatorname{char, adv}}$ head shares the hidden state $s_t$ at every time step to perform adversarial training. They are detailed as follows. Although the \textsl{attention mechanism} is not our major contribution, we include its details to facilitate reading and reference.

First, the $D_{\operatorname{char, content}}$ localizes every character by minimizing the character classification loss
\begin{equation}
\min \limits_{D_{\operatorname{char, content}}} \mathcal{L}_{\operatorname{char,content}} = - \sum_{t=1}^{| Y_{\operatorname{real}} |}{\log p(Y_{\operatorname{real}, t} \left| \right. I_{\operatorname{real}})},
\end{equation}
where the $I_{\operatorname{real}}$ denotes the real handwriting image and the $Y_{\operatorname{real}}$ denotes its corresponding text content. The $p({Y_{\operatorname{real}, t}} \left| \right. I_{\operatorname{real}})$ is the predicted conditional probability of the $t$-th ground-truth character of $I_{\operatorname{real}}$. The probability distribution of multiple character categories at the $t$-th step is obtained by 
\begin{equation}
y_t = \operatorname{Softmax}({W}_{y} {s}_{t}).
\end{equation}
The $s_t$ is the $t$-th hidden state. As suggested by the previous study~\cite{bahdanau2014neural}, it is updated by an RNN to model the context relationship within the sequence. We adopt a Gated Recurrent Unit (GRU)~\cite{cho2014learning} to obtain it as
\begin{equation}
s_t = \operatorname{GRU}\big(s_{t-1}, (Y_{\operatorname{real}, t-1}, \operatorname{feat}_{t})\big),
\end{equation}
where $\operatorname{feat}_{t}$ represents the weighted sum of the feature maps ${h}$ at location $(i, j)$ as
\begin{equation}
\operatorname{feat}_{t} = \sum_{i}\sum_{j}(\alpha_{t,ij} \ {h}_{ij}),
\end{equation}
where the ${h}$ is the feature maps extracted from the input image $I$ as
\begin{equation}
h = \operatorname{encode}(I).
\end{equation}

The vector ${\alpha}_{t}$ is the vector of attention weight, updated as follows:
\begin{equation}
\alpha_{t,ij} = \frac{\exp(e_{t,ij})}{\sum_{i,j}\big(\exp(e_{t,ij})\big)},
\end{equation}
\begin{equation}
e_{t,ij} = {W}_{e}\operatorname{Tanh}({W}_{s} \ {s}_{t-1}+{W}_{h} \ {h}_{ij}),
\end{equation}
where ${W}_{y}$, ${W}_{e}$, ${W}_{s}$ and ${W}_{h}$ are trainable parameters.

Subsequently, the separated character discriminator is able to localize and extract single characters for further adversarial training. We design another linear layer head on the top of the attention decoder with one unit output at the $t$-th step as
\begin{equation}
D_{\operatorname{char,adv}}(I, t) = {W}_{\operatorname{adv}} {s}_{t}.
\end{equation}
where the ${W}_{\operatorname{adv}}$ is a trainable parameter.
This head shares the same attention masks learned from the classification task. The adversarial loss of the $t$-th character on the generated image $I_{\operatorname{fake}}$ is formulated as follows:
\begin{equation}
\begin{split}
\min \limits_{D_{\operatorname{char, adv}}}  \mathcal{L}_{\operatorname{char, adv}}
& = \lambda \mathbb{E} \big[D_{\operatorname{char, adv}}(I_{\operatorname{fake}}, t)^2\big] \\
+ \lambda & \mathbb{E} \big[\big(1-D_{\operatorname{char, adv}}(I_{\operatorname{real}}, t)\big)^2\big], \\
\min \limits_{G} \mathcal{L}_{\operatorname{char, adv}}  = \lambda & \mathbb{E} \big[\big(1-D_{\operatorname{char, adv}}(I_{\operatorname{fake}}, t)\big)^2\big],\\
\end{split}
\end{equation}
where the hyper-parameter $\lambda$ is set to 0.1.

In addition to estimating the style-transfer effect, another objective is to retain the text content. Therefore, the discriminator also supervises the content of the generated image. With respect to the text content, the $D_{\operatorname{char, content}}$ learns from the set $\{I_{\operatorname{real}}, Y_{\operatorname{real}}\}$ and guides the generator on the set $\{I_{\operatorname{fake}}, Y_{\operatorname{print}}\}$. Note that $Y_{\operatorname{real}}$ and $Y_{\operatorname{print}}$ can be different. The optimization involves minimizing the negative log-likelihood of the conditional probability as 
\begin{equation}
\min \limits_{G} \mathcal{L}_{\operatorname{char,content}} = - \sum_{t=1}^{| Y_{\operatorname{print}} |}{\log p(Y_{\operatorname{print}, t} \left| \right. I_{\operatorname{fake}})}.
\end{equation}

\textsl{2) Cursive Join Discriminator (Adversarial Training and Handwriting Style Supervision):} Compared with the printed style image, one distinctive feature of the handwriting style image is the cursive joins between the adjacent characters. Thus, a more global discriminator (as opposed to the local discriminator for separated characters) is necessary to model the relationship between adjacent characters. Inspired by the \textsl{PatchGAN}~\cite{Isola2017patchGAN}, we use the divided patches of feature maps with overlapping receptive fields to expand the focused regions to cover adjacent characters. The discriminator (denoted as $D_{\operatorname{join}}$, consisting of two heads, termed $D_{\operatorname{join, adv}}$ and $D_{\operatorname{join, ID}}$) performs adversarial training and handwriting style supervision on these patches. The adversarial loss at the cursive join level is formulated as follows:
\begin{equation}
\begin{split}
\min \limits_{D_{\operatorname{join, adv}}}  \mathcal{L}_{\operatorname{join, adv}}
& = \mathbb{E} \big[D_{\operatorname{join, adv}}(I_{\operatorname{fake}})^2\big] \\
& + \mathbb{E} \big[\big(1-D_{\operatorname{join, adv}}(I_{\operatorname{real}})\big)^2\big], \\
\min \limits_{G} \mathcal{L}_{\operatorname{join, adv}}  & = \mathbb{E} \big[\big(1-D_{\operatorname{join, adv}}(I_{\operatorname{fake}})\big)^2\big].\\
\end{split}
\end{equation}

The feature map patches, which contain adjacent characters and the cursive joins between them, informatively indicate a specific handwriting style. Thus, this discriminator also estimates the imitation of specific handwriting styles. The $D_{\operatorname{join, ID}}$ learns from the set $\{I_{\operatorname{real}}, \operatorname{ID}\}$ and guides the style bank and the generator on the set $\{I_{\operatorname{fake}}, \operatorname{ID}\}$ by minimizing the negative log-likelihood of the conditional probability as follows:
\begin{equation}
\begin{split}
& \min \limits_{D_{\operatorname{join, ID}}} \mathcal{L}_{\operatorname{join, ID}} = - \log p(\operatorname{ID} \left| \right. I_{\operatorname{real}}), \\
& \min \limits_{\operatorname{StyleBank}, G} \mathcal{L}_{\operatorname{join, ID}} = - \log p(\operatorname{ID} \left| \right. I_{\operatorname{fake}}).
\end{split}
\end{equation}
Note that the overall framework is differentiable, thus, it can back-propagate the gradients for updating the style latent vector $\pmb{z}$ in the lookup table, namely, the style bank.

\textsl{3) Auto-encoder:} It has been observed that the GANs based on gradient descent optimization may not converge, which suggests the instability of the generator~\cite{heusel2017gans,mescheder2018training,zhang2019self,karras2020analyzing}. To this end, we set up an auto-encoder restriction of identical mapping of both the same handwriting style and content by minimizing
\begin{equation}
\mathcal{L}_{\operatorname{idt}} = \big[ G(I_{\operatorname{real}}, \pmb{z}) - I_{\operatorname{real}} \big]^2.
\end{equation}

We find that this additional design significantly contributes to the success and stability of the training, which we demonstrate in the \textsl{Experiments} section.

\begin{algorithm}[t]
\caption{Training scheme. The overall differentiable framework can back-propagate the gradients to update the style latent vector $\pmb{z}$ in the style bank.}
\label{alg-training-scheme}
\setstretch{1.2}
\KwIn{Printed style image $I_{\operatorname{print}}$ and text content $Y_{\operatorname{print}}$; Real handwriting image $I_{\operatorname{real}}$, text content $Y_{\operatorname{real}}$ and writer $\operatorname{ID}$.}
\KwOut{Updated generator $G$, style bank and dual discriminators ($D_{\operatorname{char}}$ and $D_{\operatorname{join}}$).}
\textbf{while} not converge \textbf{do}
\begin{enumerate}
\setstretch{1.8}
\item Update $D_{\operatorname{char}}$ and $D_{\operatorname{join}}$: \\
$\min \limits_{D_{\operatorname{char}}}  (\mathcal{L}_{\operatorname{char, adv}} + \mathcal{L}_{\operatorname{char, content}})$, \\
$\min \limits_{D_{\operatorname{join}}}  (\mathcal{L}_{\operatorname{join, adv}} + \mathcal{L}_{\operatorname{join, ID}})$.
\item Update $G$ and the style bank: \\
$\min \limits_{G} (\mathcal{L}_{\operatorname{char, adv}} + \mathcal{L}_{\operatorname{char, content}} + \mathcal{L}_{\operatorname{join, adv}})$, \\
$\min \limits_{\operatorname{StyleBank}, G} (\mathcal{L}_{\operatorname{join, ID}} + \mathcal{L}_{\operatorname{idt}})$.
\end{enumerate}
\end{algorithm}

\textsl{4) Training Scheme:} The training scheme of the proposed method is shown in Algorithm~\ref{alg-training-scheme}. The generator and discriminators are updated in alternating steps to achieve adversarial training. We parameterize the handwriting styles by jointly training the generator and the style bank.

\begin{table}[h]
\scriptsize
\centering
\caption{Architecture of the generator. The k, s, p represent kernel, stride and padding sizes, respectively. For instance, k:3 represents a $3\times3$ kernel size. ``N.", ``R." and ``Tanh" stand for batch normalization, ReLU and Tanh layer, respectively.}
\label{table:architecture-generator}
\setlength{\tabcolsep}{1.mm}{
\begin{tabular}{ccc}
\toprule
Type & Configurations  & Out Size \\
\midrule
Input & - & 3$\times$64$\times$400 \\
\midrule
Conv  & $\mathrm{maps:16, k:5, s:1, p:2, N.+R.}$ & $16 \times 64 \times 400$ \\
Conv  & $\mathrm{maps:32, k:3, s:2, p:1, N.+R.}$ & $32 \times 32 \times 200$ \\
Conv  & $\mathrm{maps:64, k:3, s:2, p:1, N.+R.}$ & $64 \times 16 \times 100$ \\
Conv  & $\mathrm{maps:128, k:3, s:2, p:1, N.+R.}$ & $128 \times 8 \times 50$ \\
Conv  & $\mathrm{maps:256, k:3, s:2, p:1, N.+R.}$ & $256 \times 4 \times 25$ \\
\midrule
Res. Block & $ \begin{bmatrix} \mathrm{maps:256, k:3, s:1, p:1, N.+R.} \\ \mathrm{maps:256, k:3, s:1, p:1, N.} \end{bmatrix} *6 $ & 256$\times$4$\times$25 \\
\midrule
Deconv  & $\mathrm{maps:128, k:3, s:2, p:1, N.+R.}$ & $128\times8\times50$ \\
Deconv  & $\mathrm{maps:64, k:3, s:2, p:1, N.+R.}$ & $64\times16\times100$ \\
Deconv  & $\mathrm{maps:32, k:3, s:2, p:1, N.+R.}$ & $32\times32\times200$ \\
Deconv  & $\mathrm{maps:16, k:s3, s:2, p:1, N.+R.}$ & $16\times64\times400$ \\
Deconv  & $\mathrm{maps:3, k:5, s:2, p:2, Tanh}$ & $3\times64\times400$ \\
\bottomrule
\end{tabular}
}
\end{table}

\subsection{Inference of Generator}

After the training, it is possible to generate new handwriting styles by manipulating the elements of the latent vector $\pmb{z} \in \mathbb{R}^d$, namely, $\pmb{z} = \{z_1, ..., z_k, ..., z_d\}$. To create a random handwriting style, we adjust every $z_k$ in the range of 
\begin{equation}
[\operatorname{min}(\pmb{z}_{all, k}), \operatorname{max}(\pmb{z}_{all, k})],
\end{equation}
where $\pmb{z}_{all} \in \mathbb{R}^{d \times n}$ stores all the $\pmb{z}$ embedded from $n$ writer IDs in the training set. The restriction of the range of $z_k$ avoids the failed generation, because the limited $z_k$ falls into the modes that the generator ever seen.

Moreover, the input printed style image, which serves as a text content condition, can be changed to achieve different effects, including different position arrangements (for instance, to generate curve text) and text of arbitrary length (for instance, to generate a sentence).

\begin{table}[t]
\scriptsize
\centering
\caption{Architecture of the separated character discriminator. Each convolutional layer with a kernel size of 3, a stride of 1 and a padding size of 1. ``N.", ``R." and ``AvgPool" stand for instance normalization, PReLU and average pooling layer, respectively.}
\label{table:architecture-discriminator-sep}
\setlength{\tabcolsep}{1.mm}{
\begin{tabular}{ccc}
\toprule
Type & Configurations  & Out Size \\
\midrule
Input & - & 3$\times$64$\times$400 \\
\midrule
Conv  & $\mathrm{maps:16, N.+R.+AvgPool:2\times2}$ & $16 \times 32 \times 200$ \\
Conv  & $\mathrm{maps:64, N.+R.+AvgPool:2\times2}$ & $64 \times 16 \times 100$ \\
Conv  & $\mathrm{maps:128, N.+R.+AvgPool:2\times2}$ & $128 \times 8 \times 50$ \\
Conv  & $\mathrm{maps:128, N.+R.+AvgPool:2\times2}$ & $128 \times 4 \times 25$ \\
Conv  & $\mathrm{maps:192, N.+R.+AvgPool:2\times1}$ & $192 \times 2 \times 25$ \\
Conv  & $\mathrm{maps:256, N.+R.}$ & $256 \times 2 \times 25$ \\
Conv  & $\mathrm{maps:256, N.+R.}$ & $256 \times 2 \times 25$ \\
\midrule
Att. & $ \mathrm{256\ hidden\ units, 256\ GRU\ units} $ & - \\
\midrule
2 FC  & $\mathrm{adversary: 1\ unit, classifier: \ } n\ \mathrm{characters}$ & - \\
\bottomrule
\end{tabular}
}
\end{table}

\begin{table}[t]
\scriptsize
\centering
\caption{Architecture of the cursive joint discriminator, which shares the first four convolutions with the separated character discriminator. Each convolutional layer has a kernel size of 3, a stride of 1 and a padding size of 1. ``N.", ``R." and ``AvgPool" stand for instance normalization, PReLU and average pooling layer, respectively.}
\label{table:architecture-discriminator-cur}
\setlength{\tabcolsep}{1.mm}{
\begin{tabular}{cccc}
\toprule
Head & Type & Configurations  & Out Size \\
\midrule
- & Input & - & 3$\times$64$\times$400 \\
\midrule
\multirow{4}{*}{-} & Conv  & $\mathrm{maps:16, N.+R.+AvgPool:2\times2}$ & $16 \times 32 \times 200$ \\
& Conv  & $\mathrm{maps:64, N.+R.+AvgPool:2\times2}$ & $64 \times 16 \times 100$ \\
& Conv  & $\mathrm{maps:128, N.+R.+AvgPool:2\times2}$ & $128 \times 8 \times 50$ \\
& Conv  & $\mathrm{maps:128, N.+R.+AvgPool:2\times2}$ & $128 \times 4 \times 25$ \\
\midrule
\multirow{3}{*}{Adversary} & Conv  & $\mathrm{maps:64, N.+R.+AvgPool:2\times2}$ & $64 \times 2 \times 13$ \\
& Conv  & $\mathrm{maps:16, N.+R.}$ & $16 \times 2 \times 13$ \\
& Conv  & $\mathrm{maps:1}$ & $1 \times 2 \times 13$ \\
\midrule
Style & Conv  & $\mathrm{maps:192, N.+R.+AvgPool:2\times2}$ & $192 \times 2 \times 13$ \\
Classifier & Conv  & $\mathrm{maps:256, N.+R.+AvgPool:2\times2}$ & $256 \times 1 \times 7$ \\
& Conv  & $\mathrm{maps: \ } n\ \mathrm{styles, AvgPool:1\times7}$ & $n$ \\
\bottomrule
\end{tabular}
}
\end{table}

\section{Experiments}

\subsection{Datasets}

\textbf{IAM}~\cite{marti2002iam} contains more than 13,000 lines and 115,000 words written in English by 657 different writers. It can serve as a basis for a variety of handwriting recognition tasks.

\textbf{RIMES}~\cite{augustin2006rimes} contains more than 60,000 words written in French by over 1,000 authors. Its goal is to evaluate a system dedicated to handwriting recognition and indexing documents.

\textbf{CVL}~\cite{Kleber2013cvl} contains seven different handwritten texts (one in German and six in English) written by 311 different writers. We use the English part for the experiment of domain adaptation.

In addition, we use the default training subsets and testing subsets of the above datasets for the following experiments.

\subsection{Implementation Details}

\begin{table*}[h]
\centering
\caption{Ablation study. The metrics are FID and GS. Lower values are preferable.}
\label{table:ablation-study}
\setlength{\tabcolsep}{2.5mm}{
\begin{tabular}{ c c c c c c c c}
\toprule
$\mathcal{L}_{\operatorname{join, adv}}$ & $\mathcal{L}_{\operatorname{idt}}$ & $\mathcal{L}_{\operatorname{char, content}}$ & $\mathcal{L}_{\operatorname{char, adv}}$ & $\mathcal{L}_{\operatorname{join, ID}}$ & FID & GS & Image (``tomorrow") \\
\midrule
\checkmark & & & & & 272.80 & $ 5.00 \times 10^{-2}$ & \includegraphics[width=0.25\columnwidth]{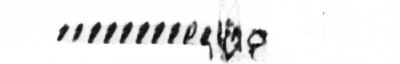} \\
\checkmark & \checkmark & & & & 48.02 & $ 6.84 \times 10^{-3}$ & \includegraphics[width=0.25\columnwidth]{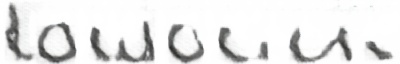} \\
\checkmark & \checkmark & \checkmark & & & 15.68 & $ 8.59 \times 10^{-4}$ & \includegraphics[width=0.25\columnwidth]{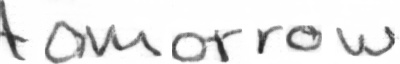}\\
\checkmark & \checkmark & \checkmark & \checkmark & & 14.20 & $ 7.82 \times 10^{-4}$ & \includegraphics[width=0.25\columnwidth]{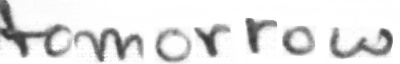} \\
\checkmark & \checkmark & \checkmark & \checkmark & \checkmark & \textbf{12.06} & $\pmb{5.59 \times 10^{-4}}$ & \includegraphics[width=0.25\columnwidth]{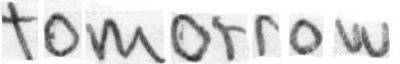}\\
\bottomrule
\end{tabular}
}
\end{table*}

\textbf{Network} Our generator consists of five convolutions, six residual blocks~\cite{he2016deep} and five deconvolutions, as shown in Table~\ref{table:architecture-generator}. The style vector, whose dimensionality is set to 256, is concatenated with the output feature maps of the third residual block. 

The dual discriminators shares first four convolutions, whose architectures are shown in Table~\ref{table:architecture-discriminator-sep} and~\ref{table:architecture-discriminator-cur}. 

\textbf{Optimization} We use the ADAM~\cite{kingma2014adam} as the optimizer with the settings of $\beta_1 = 0.5$ and $\beta_2 = 0.999$. The learning rate is set to $10^{-4}$ and linearly decreased to $10^{-5}$ after 300,000 iterations. The batch size is 128. The images are resized to a height of 64 pixels, maintaining the ratio. If the image width is less than 400, we pad the image with white to obtain a width of 400 pixels. Otherwise, we resize it to 400. Our method is built on the PyTorch framework~\cite{pytorch} and all experiments are conducted on NVIDIA 1080Ti GPUs.

\subsection{Evaluation Metrics}

Following the settings of~\cite{Fogel2020ScrabbleGAN,Alonso2019Adversarial}, the evaluation metrics are the widely used GAN metrics of Fr\'echet Inception Distance (FID)~\cite{heusel2017gans} and Geometric-Score (GS)~\cite{khrulkov2018geometry}. The FID captures the similarity of generated images to real ones, which is validated to be consistent with human judgment. The GS compares the geometrical properties of the underlying data manifold and the generated one, which provides both qualitative and quantitative means for evaluation. Lower values of FID and GS are preferable.

With respect to recognition performance, we use the Word Error Rate (WER) and Character Error Rate (CER) as metrics for handwritten text recognition. The WER denotes the ratio of the mistakes at the word level, among all the words of the ground truth, and the CER measures the Levenshtein distance normalized by the length of the ground truth. Lower values of WER and CER are preferable.

\subsection{Ablation Study}

We study the effectiveness of the proposed components on RIMES and list five results in a progressive combination manner in Table~\ref{table:ablation-study}. First we build a baseline using only $\mathcal{L}_{\operatorname{join, adv}}$. Thus, the framework degenerates to a  \textsl{PatchGAN}~\cite{Isola2017patchGAN} using only an adversarial loss. The latent vector $\pmb{z}$ is replaced by a noise vector $\pmb{n}\sim \mathcal{N}(0, 1)$. As illustrated by the images in Table~\ref{table:ablation-study}, the baseline generation contains meaningless strokes and falls into a collapse mode. The FID and GS scores are large, which indicates a large gap between the source and the generated images.

Then we add the identical mapping loss $\mathcal{L}_{\operatorname{idt}}$ to ease the training. We find that the auto-encoder significantly contributes to the success and stability of the training. Some glyphs can be seen in the generated images. To further guide the generator to retain the text content on the output image, we add a content supervision using the character-level content loss $\mathcal{L}_{\operatorname{char, content}}$. The word ``tomorrow" can be distinguished from the generated image. Subsequently, we add a character-level adversarial loss $\mathcal{L}_{\operatorname{char, adv}}$ to improve the reality of every character. The generated image is quite realistic and satisfactory. This indicates that the dual discriminators for two-level style adversarial learning are critical to the high-quality of the generated handwritten text image. Furthermore, we require the generator to present a specific handwriting style that can be easily identified. Under the supervision of the writer IDs, the handwriting styles are parameterized as corresponding latent vectors stored in the style bank, which serve as style priors for the generator. By this way, more handwriting features occur, including character slant, cursive join, stroke width, ink blot, paper backgrounds, \textsl{etc}. We obtain the best FID of $12.06$ and GS of $5.59 \times 10^{-4}$.

\subsection{Comparison with Previous Methods}

\begin{table}[t]
\centering
\caption{Comparison with previous methods.}
\label{table:GAN-metric}
\setlength{\tabcolsep}{5mm}{
\begin{tabular}{ c c c }
\toprule
Method & FID & GS \\
\midrule
Alonso \textit{et al.}~\cite{Alonso2019Adversarial} & 23.94 & $ 8.58 \times 10^{-4}$ \\
ScrabbleGAN~\cite{Fogel2020ScrabbleGAN} & 23.78 & $ 7.60 \times 10^{-4}$ \\
HiGAN~\cite{gan2021higan} & 17.28 & - \\
SLOGAN (Ours) & \textbf{12.06} & $\pmb{5.59 \times 10^{-4}}$ \\
\bottomrule
\end{tabular}
}
\end{table}

\begin{figure}[t] 
\centering
\includegraphics[width=0.95\columnwidth]{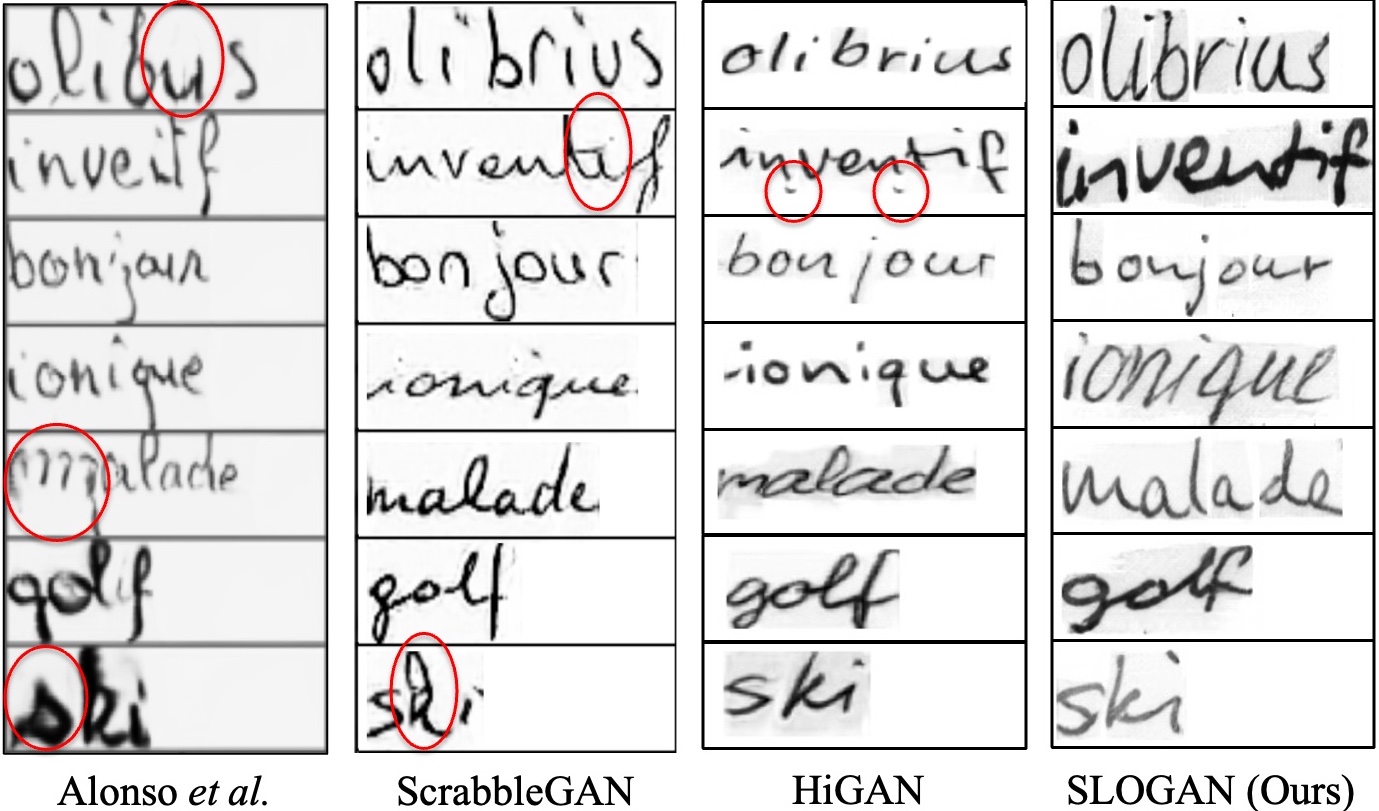} 
\caption{Visual comparison with previous studies. From left to right: images from Alonso \textit{et al.}~\cite{Alonso2019Adversarial}, ScrabbleGAN~\cite{Fogel2020ScrabbleGAN}, HiGAN~\cite{gan2021higan}, and our SLOGAN. The words in each row are the same. The unclear characters and unexpected artifacts are indicated by red circles. The proposed method can generate clearer characters with fewer unexpected artifacts.}
\label{pic-comp-prev}
\end{figure}

Applying settings similar to those in~\cite{Fogel2020ScrabbleGAN,Alonso2019Adversarial}, we evaluate the quality of our generated images under the metric of FID (using 25k real and 25k generated images) and GS (using 5k real and 5k generated images). Note that our evaluation is more rigorous, as recommended in~\cite{Fogel2020ScrabbleGAN}. We finish the training before evaluation, as opposed to on-the-fly evaluation during training and the choice of the best score~\cite{Alonso2019Adversarial}. 

\begin{table}[t]
\centering
\caption{Comparison on out-of-vocabulary word images. }
\label{table:OOV}
\setlength{\tabcolsep}{10mm}{
\begin{tabular}{ c c }
\toprule
Method & FID \\
\midrule
GANwriting~\cite{Kang2020GANwriting} & 125.87 \\
SLOGAN (Ours) & \textbf{97.81} \\
\bottomrule
\end{tabular}
}
\end{table}

\begin{figure}[t] 
\centering
\includegraphics[width=0.95\columnwidth]{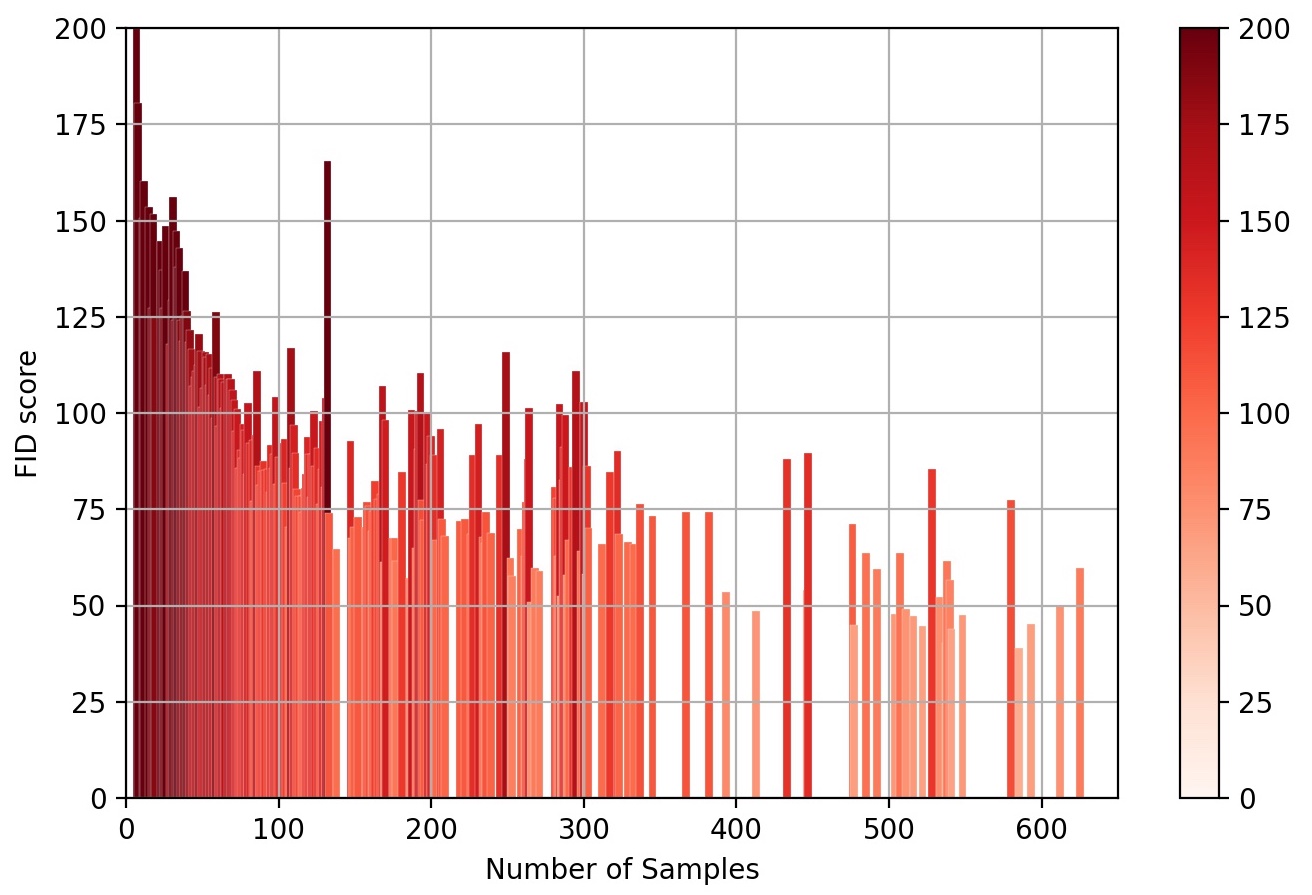} 
\caption{FID scores for out-of-vocabulary word images in existing handwriting styles. These styles are sorted by the number of training samples. Lower values of FID are preferable.}
\label{pic-oov-bar}
\end{figure}

\begin{figure*}[t] 
\centering
\includegraphics[width=1.8\columnwidth,height=0.94\columnwidth]{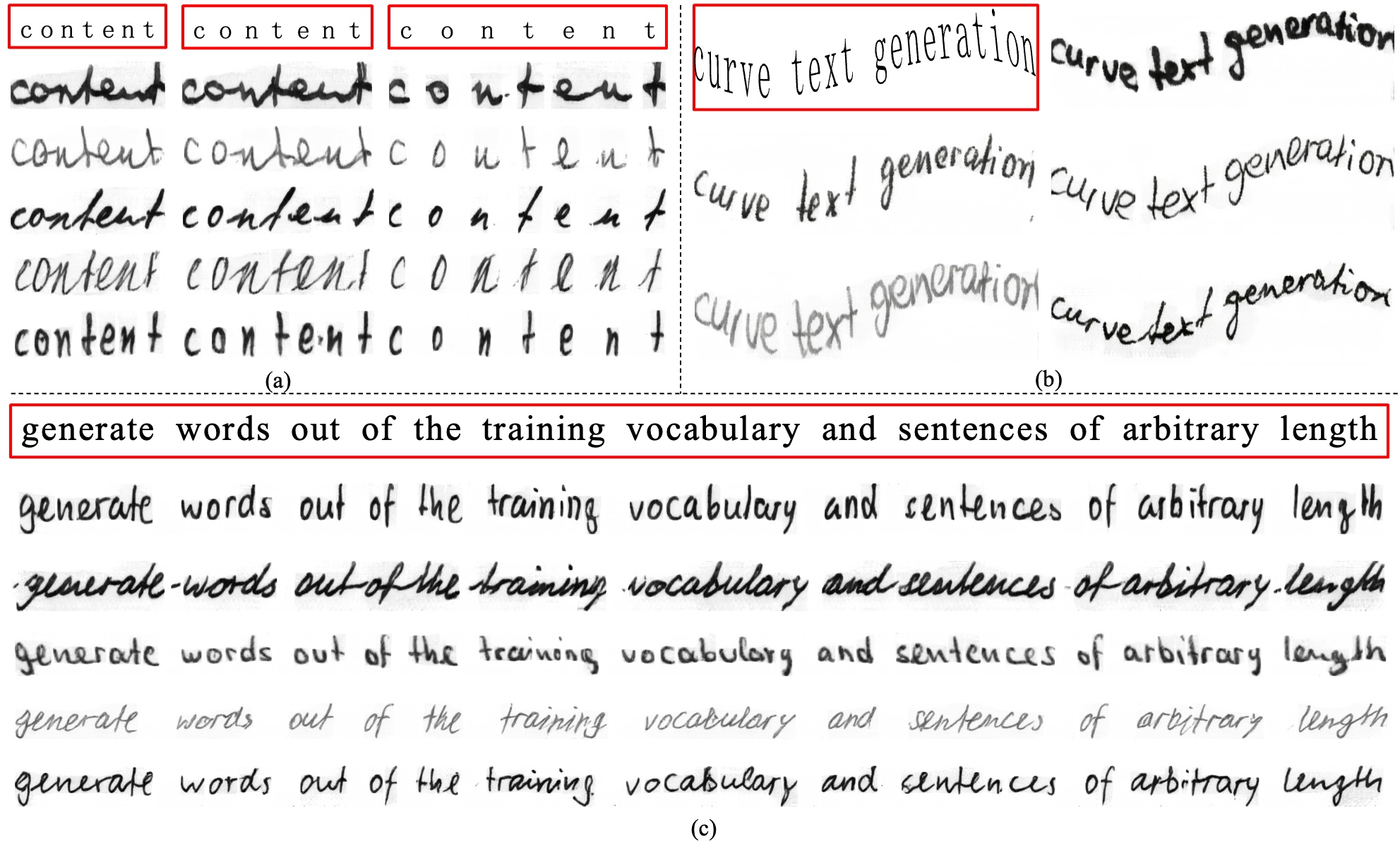} 
\caption{Content diversity that previous studies~\cite{Fogel2020ScrabbleGAN,Alonso2019Adversarial,Kang2020GANwriting,gan2021higan} cannot achieve. Effects include (a) adjacent character interval, (b) curved text and (c) arbitrary length sentence with reasonably space between words, produced by variant input images, which are indicated by red boxes. We illustrate five styles of generated samples for every input image. Zoom in for better view.}
\label{pic-condition-content}
\end{figure*}

\begin{figure*}[t] 
\centering
\includegraphics[width=1.\columnwidth, height=0.8\columnwidth]{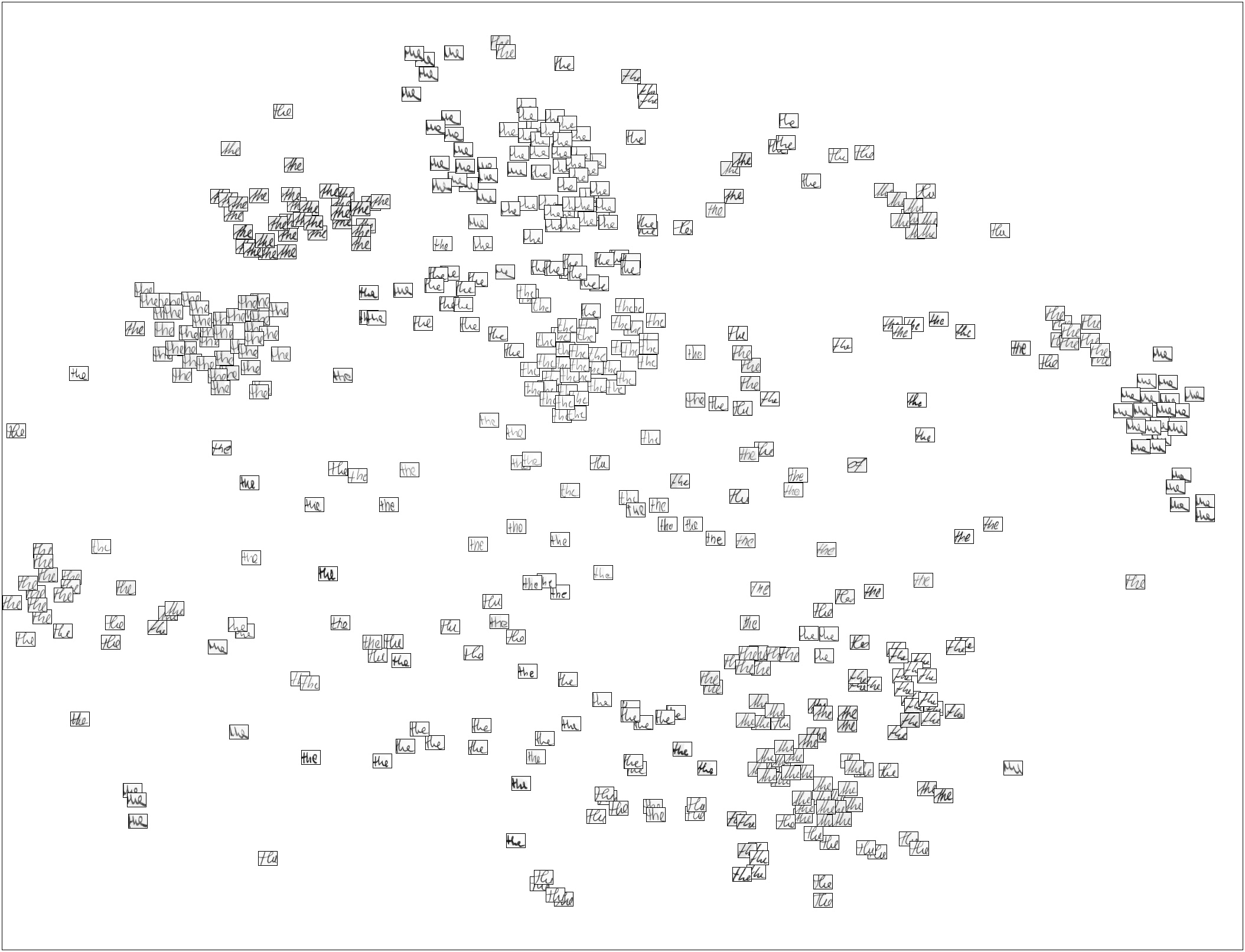} 
\includegraphics[width=1.\columnwidth, height=0.8\columnwidth]{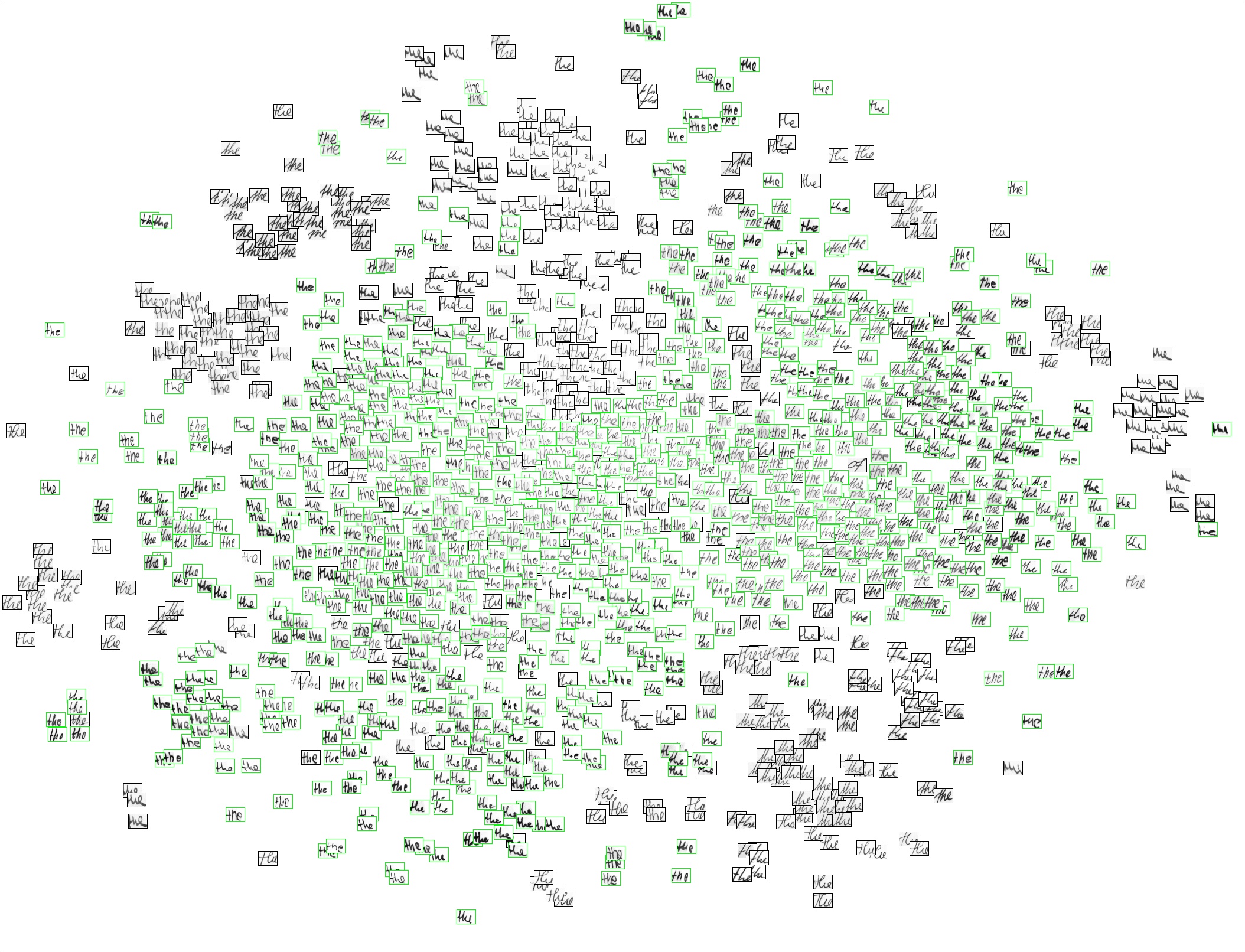}
\caption{Distribution of the handwriting styles of the word ``the" via t-SNE. Left: distribution of the existing styles in the IAM training set~\cite{marti2002iam}; right: distribution after adding our generated samples (with the same content ``the", marked by green bounding boxes) to the existing training set. The large amount of empty space in the original distribution suggests the limitation of styles. With our generated various styles, the distribution is more even and reasonable, which indicates the bias of the style is significantly rectified. Zoom in for better view.}
\label{pic-tSNE-the}
\end{figure*}

As presented in Table~\ref{table:GAN-metric}, the proposed SLOGAN achieves best results on both FID and GS metrics. The SLOGAN gains benefit from the more auxiliary objectives and more fine-grained supervisions, including two level guidances, content and writer ID supervisions. As illustrated in Figure~\ref{pic-comp-prev}, we generate clearer characters with fewer unexpected artifacts.

\subsection{Generating Out-of-Vocabulary Words}

By changing only the input image, the SLOGAN is able to synthesize words out of the training vocabulary. To validate the effectiveness and generality of our method, we conduct an experiment to generate out-of-vocabulary words following a more stringent regulation of GANwriting~\cite{Kang2020GANwriting}, in which the final FID score is the average of the FID scores of each handwriting style, rather than directly performing evaluation on randomly sampled generated images. Under this setting, a handwriting style with only a few training samples takes the same proportion as a style with hundreds of samples. This means our method is required to perform well despite the few-shot learning of certain handwriting styles. As shown in Figure~\ref{pic-oov-bar}, the FID score decreases when the number of the training samples increases. We find that our method can significantly imitate a handwriting style by learning from approximately 20 samples.

\begin{figure*}[t] 
\centering
\includegraphics[width=1.\columnwidth, height=0.85\columnwidth]{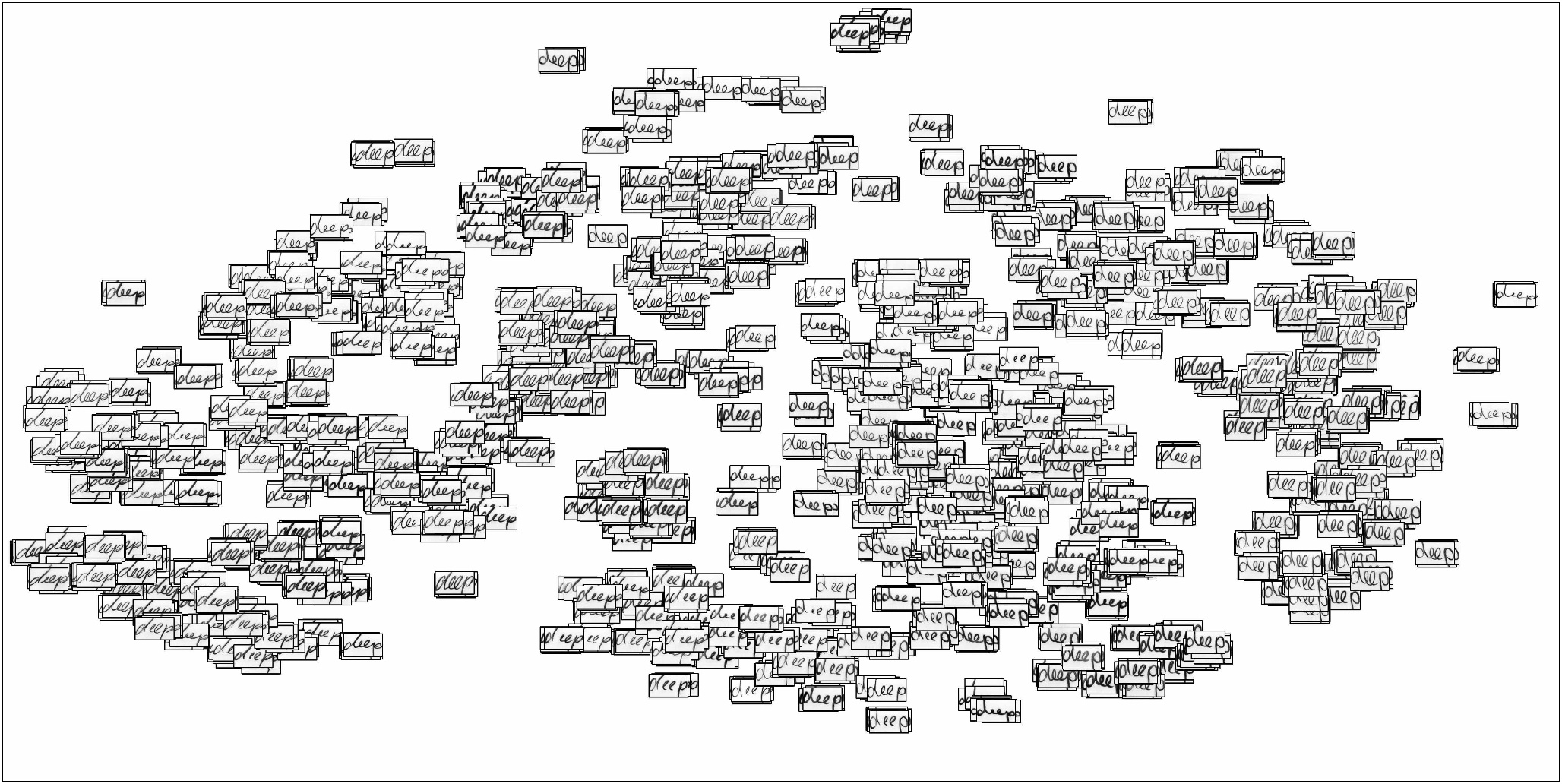}
\includegraphics[width=1.\columnwidth, height=0.85\columnwidth]{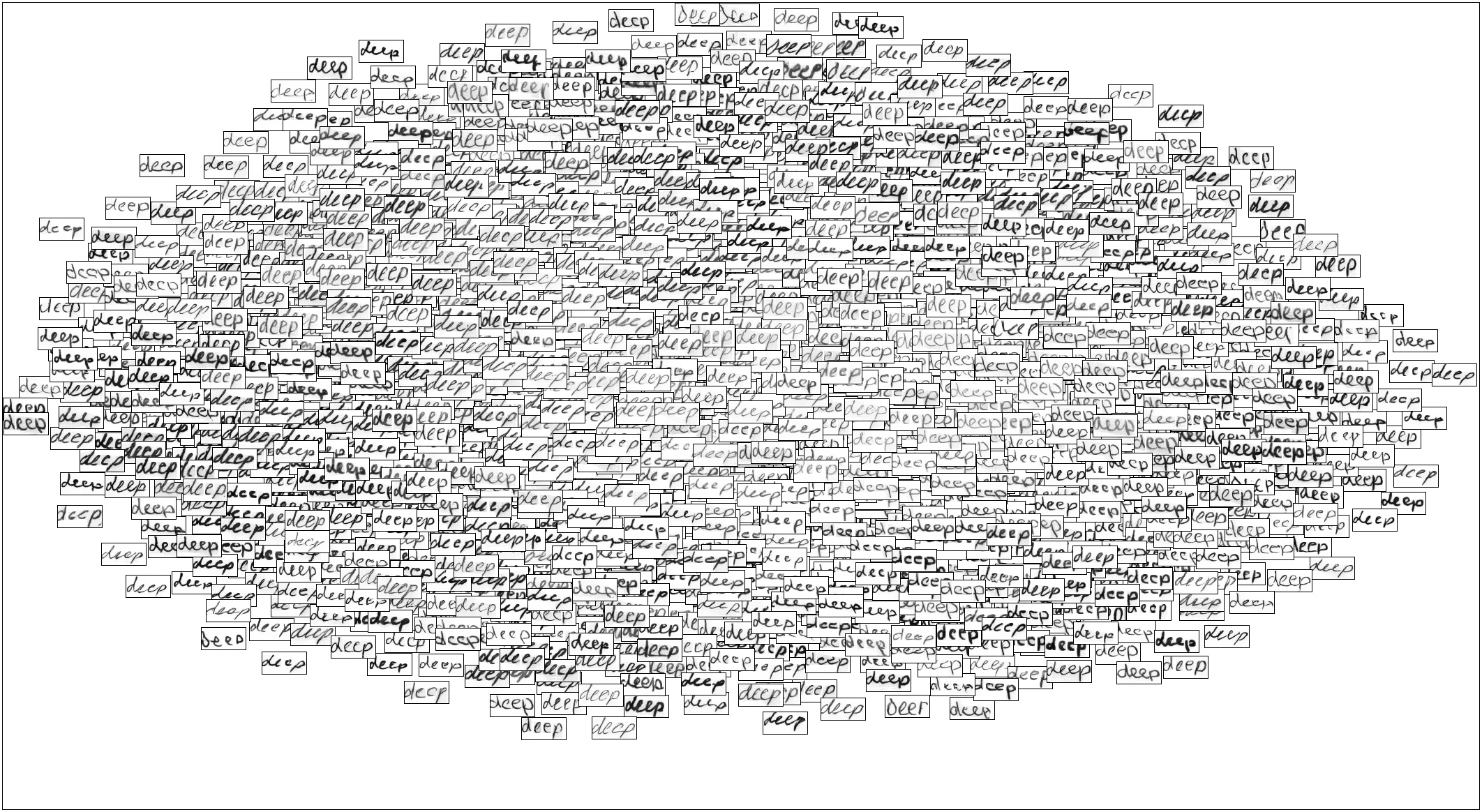}
\caption{Distribution of the handwriting styles of the word ``deep" via t-SNE. Left: distribution of samples generated by GANwriting~\cite{Kang2020GANwriting}; right: our distribution. Following the setting of Kang \textit{et al.}~\cite{Kang2020GANwriting}, we generate 2,500 samples for t-SNE embedding. Our sample distribution is more even and reasonable, which suggests the superior of the diversity of our method. Zoom in for better view.}
\label{pic-tSNE-deep}
\end{figure*}

We choose the corresponding result (with existing style but unseen content) of GANwriting~\cite{Kang2020GANwriting} for a fair comparison, because the generated images containing out-of-vocabulary words are bounded by existing handwriting styles in the training set. As shown in Table~\ref{table:OOV}, although the previous study presented promising generated images, our method takes a further step and outperforms it by a notable margin. 

\subsection{Diversity of Generation}

The generator learns to generate images according to the input printed style image $I_{\operatorname{print}}$ and the latent style vector $\pmb{z}$. Once we change the $I_{\operatorname{print}}$ and $\pmb{z}$, the generated image achieves different effects, which increases the diversity of the training data. 

\begin{figure}[t] 
\centering
\includegraphics[width=1\columnwidth]{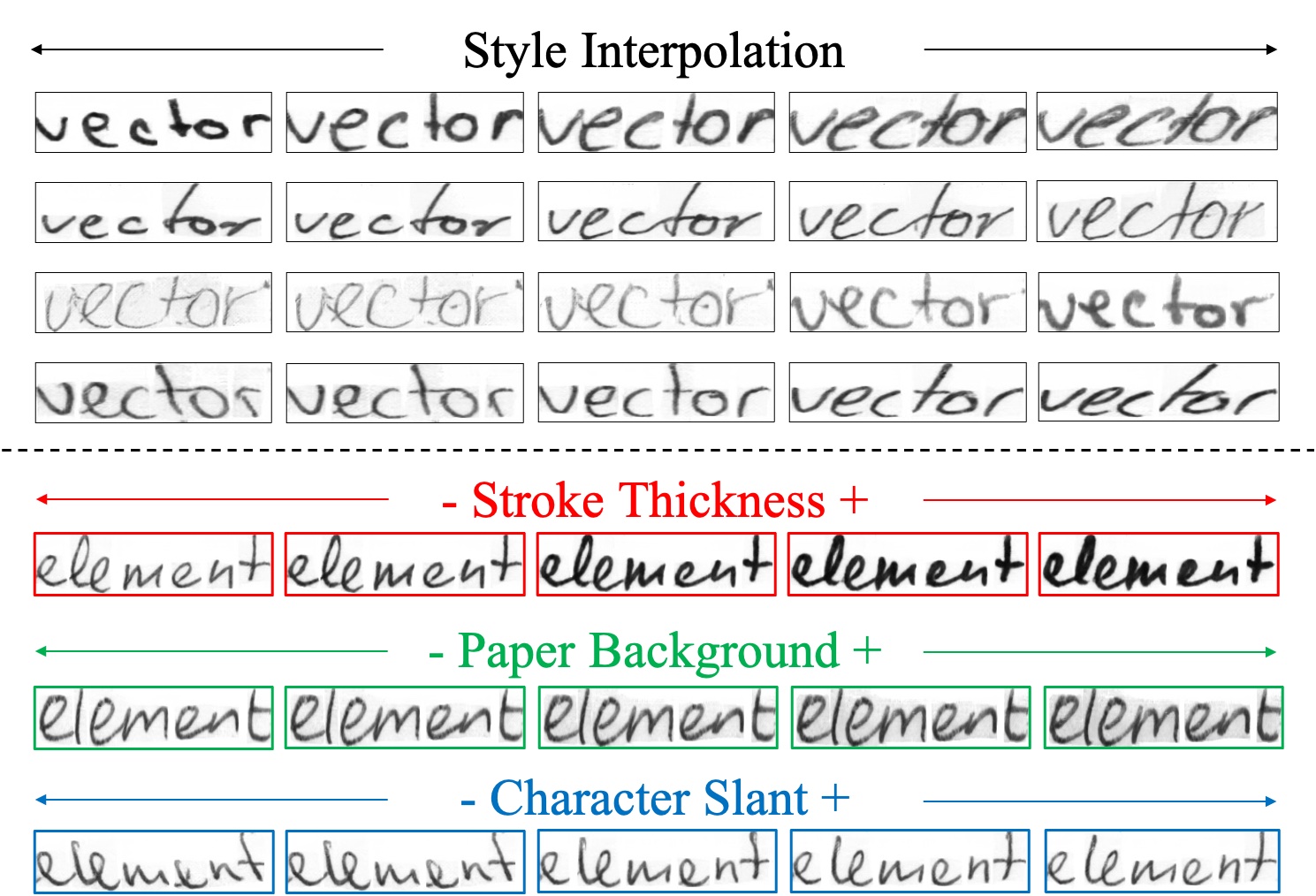} 
\caption{Style diversity produced by latent vector interpolation (top) and certain element manipulations (bottom).}
\label{pic-condition-style}
\end{figure}

\textbf{Content Diversity} Because the generator works in a fully convolutional fashion, there exists spatial consistency between the input and output images. As shown in Figure~\ref{pic-condition-content}, we present the different effects, including adjacent character interval, curved text and arbitrary length text, produced by the variant input images. 

For instance, we expand the interval between adjacent characters on the input image and the character interval on the generated image is enlarged as well. Moreover, using a curved printed text image as input, the generator also outputs a curved handwritten text image. Simultaneously, the generator trained on word-level images is able to synthesize sentences of arbitrary length, requiring only a change of sentence string in the input printed style image. The proposed SLOGAN can reasonably tackle the space between words, whereas previous studies~\cite{Fogel2020ScrabbleGAN,gan2021higan} can only generate a sentence as a long word without any space. From this perspective, the proposed SLOGAN is easy and flexible to use. 

\textbf{Style Diversity} Handwriting styles are parameterized as latent vector $\pmb{z} = \{z_1, ..., z_k, ..., z_d\}$, in which the element $z_k$ is manipulatable to control the generated styles. As shown in Figure~\ref{pic-tSNE-the}, we randomly introduce perturbations to the parameters to generate various handwriting styles. After adding the generated samples (marked by green bounding boxes) to the existing training set, the distribution is more even and reasonable, which indicates that the bias of the style is significantly rectified. 

We also compare our method to the advanced GANwriting~\cite{Kang2020GANwriting}. As demonstrated in Figure~\ref{pic-tSNE-deep}, we randomly generate various styles of the word ``deep" following the setting of Kang \textit{et al.}~\cite{Kang2020GANwriting} and visualize the distribution via t-SNE. It can be seen that the distribution of our generated samples is more even, which suggests the superiority of the diversity of our data synthesis. Because the original paper did not report recognition accuracy gains, we only conduct a comparison based on the visual effects for reference.

Furthermore, we demonstrate the visual effects achieved by manipulating the latent vector in two ways. As shown in Figure~\ref{pic-condition-style}, we interpolate the entire vector $\pmb{z}$ between two random vectors $\pmb{z_a}$ and $\pmb{z_b}$ to achieve style interpolation, and to manipulate certain elements $z_k$ to achieve special attribute changes.

\subsection{Improvement of Recognition Performance}

The various synthesized data significantly contribute to the diversity of the training samples, thereby benefiting the robustness of the text recognizers. To validate the effectiveness of the generated samples and compare them with previous studies~\cite{Fogel2020ScrabbleGAN,Alonso2019Adversarial}, we follow the setting that adds 100K generated samples for training. However, previous studies used different recognizers, which are not directly comparable. Thus, we use a more common and generalized recognizer~\cite{Bhunia2019Handwriting,luo2020learn} in the subsequent experiments. As presented in Table~\ref{table:size}, the extra synthesized data significantly boost the recognition performance. For instance, the WER of IAM significantly decreases ($\downarrow$1.83\%, from 19.12\% to 17.29\%). Actually, our baseline is higher than that of~\cite{Fogel2020ScrabbleGAN}, indicating less room for improvement, but we gain more error reduction than theirs (less than 1\%). 

Another interesting observation is that the 100K extra samples are barely sufficient for training. We find that approximately 10 million synthesized samples can reach the upper bound of our method. As shown in Table~\ref{table:size}, 10 million samples further improve the recognition performance by a significant margin, especially on the WER of IAM ($\downarrow$2.32\%, from 17.29\% to 14.97\%).

\begin{table}[t]
\centering
\caption{Diverse synthesized data increases the robustness of the recognizer. The ``$+\infty$" indicates training using as many as possible synthesized data. In our practice, approximately ten million samples are sufficient for training.}
\setlength{\tabcolsep}{4.5mm}{
\begin{tabular}{ c c c c}
\toprule
Set & Synth. Data Size & WER & CER     \\ 
\midrule
\multirow{3}{*}{IAM}
& - & 19.12 & 7.39 \\
& $+$100K & 17.29 & 6.76 \\ 
& $+\infty$ ($\sim$10M) & \textbf{14.97} & \textbf{5.95} \\
\midrule
\multirow{3}{*}{RIMES}
& - & 13.83 & 3.93 \\ 
& $+$100K & 12.01 & 3.50 \\ 
& $+\infty$ ($\sim$10M) & \textbf{11.50} & \textbf{3.35} \\
\bottomrule
\end{tabular}
\label{table:size}
}
\end{table}

\begin{table}[t]
\centering
\caption{Complementariness of the data synthesis and the data augmentation. We obtain the best result using both data synthesis and data augmentation.}
\setlength{\tabcolsep}{2.5mm}{
\begin{tabular}{ c c c c c}
\toprule
\multirow{2}{*}{Method} & \multicolumn{2}{c}{IAM} & \multicolumn{2}{c}{RIMES}     \\ 
\cmidrule(lr){2-3} \cmidrule(lr){4-5}
& WER & CER & WER & CER \\
\midrule
Sueiras \textit{et al.}~\cite{sueiras2018offline} & 23.80 & 8.80 & 15.90 & 4.80 \\
Alonso \textit{et al.}~\cite{Alonso2019Adversarial} & - & - & 11.90 & 4.03 \\
Zhang \textit{et al.}~\cite{zhang2019sequence} & 22.20 & 8.50 & - & - \\
Bhunia \textit{et al.}~\cite{Bhunia2019Handwriting} & 17.19 & 8.41 & 10.47 & 6.44 \\
Fogel \textit{et al.}~\cite{Fogel2020ScrabbleGAN} & 23.61 & 13.42 & 11.32 & 3.57 \\
Kang \textit{et al.}~\cite{Kang2020Unsupervised} & 17.26 & 6.75 & - & - \\
\midrule
Baseline & 19.12 & 7.39 & 13.83 & 3.93   \\
$+$SLOGAN (Ours) & 14.97 & 5.95 & 11.50 & 3.35 \\
$+$Aug.~\cite{luo2020learn} & 14.04 & 5.34 & 9.23 & 2.57 \\
$+$SLOGAN $+$Aug. & \textbf{12.90} & \textbf{4.94} & \textbf{8.80} & \textbf{2.44} \\
\bottomrule
\end{tabular}
\label{table:SOTA}
}
\end{table}

We are also interested in the relationship between the data augmentation and the data synthesis. As discussed in the \textsl{Related Work} section, data augmentation is performed on existing samples and cannot create new images containing words out of the training vocabulary, whereas data synthesis makes these images available. Therefore, we integrate our data synthesis method with data augmentation by using the open-source toolkit\footnote{\url{https://github.com/Canjie-Luo/Text-Image-Augmentation}} and performing random augmentation~\cite{luo2020learn} on the synthesized samples. For a fair comparison, the methods listed in Table~\ref{table:SOTA} share similar training settings. Methods using additional data or language models are outside the scope of this study. As shown in Table~\ref{table:SOTA}, independent augmentation and synthesis benefit the robustness of the recognizer. Moreover, the recognizer trained using both data synthesis and data augmentation achieves the best performance, which suggests the complementariness of the data synthesis and the data augmentation. 

\subsection{Domain Adaptation}

\begin{table}[t]
\centering
\caption{Domain adaptation by using IAM training dataset and using CVL lexicon to synthesize samples. We evaluate the recognition performance on the CVL testing dataset.}
\setlength{\tabcolsep}{1.mm}{
\begin{tabular}{ c c c c }
\toprule
Method & Training Data & WER & CER     \\
\midrule
Baseline & IAM & 42.72$\pm$0.12 & 18.39$\pm$0.41  \\
\midrule
ScrabbleGAN~\cite{Fogel2020ScrabbleGAN} & IAM+GAN & 35.98$\pm$0.38 & 17.27$\pm$0.23 \\
SLOGAN (Ours) & IAM+GAN & \textbf{34.98$\pm$0.31} & \textbf{14.10$\pm$0.10} \\
\bottomrule
\end{tabular}
\label{table:adaptation}
}
\end{table}

We further explore the potential capacity for domain adaptation by training the recognizer using the IAM training set and CVL-like generated samples, and evaluating on the CVL testing set. Following the settings of ScrabbleGAN~\cite{Fogel2020ScrabbleGAN}, we train our GAN using only the IAM training dataset and generate 100K samples using the CVL lexicon as additional training data. We repeat the training five times and report the averages and standard deviations thereof. As shown in Table~\ref{table:adaptation}, the reproduced baseline is comparable to that of the original study. After adding 100K generated samples, the recognition performance is significantly boosted. The recognizer trained using our synthesized data achieves lower WER and CER, indicating the superior quality and diversity of our generated data.

\subsection{Turing Test}

We have studied whether or not the generated images are indistinguishable from the real handwriting samples by conducting two human evaluation experiments. We recruited 40 volunteers who had once engaged in text image processing tasks, including text region detection, text image recognition, text image enhancement, \textit{etc.}, because we believe the person who worked in related works owns superior acuity and insight into the variance of the text patterns. 

First, we showed 20 representative real handwriting samples to the volunteers. They were then asked to classify an image as real one or generated one. The images for classification were randomly collected from the real training images and our generated images. We collected 50 valid responses from every volunteer, that is, 2,000 responses in total. The results are presented as a confusion matrix in Table~\ref{table:human-1}, including Recall (Rec), Precision (Pre), False Positive Rate (FPR), False Omission Rate (FOR) and Accuracy (Acc) values. We find that the classification accuracy is close to 50\%, which suggests that it is almost a random binary classification. This indicates that even the expert volunteers cannot easily identify the generated images. 

\begin{table}[t]
    \caption{Confusion matrix in the first \textit{Turing Test}. The volunteers are asked to classify an image as real one or generated one. The values include Recall (Rec), Precision (Pre), False Positive Rate (FPR), False Omission Rate (FOR) and Accuracy (Acc).}
    \label{table:human-1}
    \centering
    \begin{tabular}{c c c c}
    \toprule
        \multirow{2}{*}{Actual} & \multicolumn{2}{c}{Predicted}\\
        \cmidrule{2-3}
         & Real & Fake&\\
         \midrule
         Genuine & 31.50  & 20.50 & Rec: 60.58 \\
         Generated & 28.55 & 19.45 &  FPR: 59.48 \\
        &  Pre: 52.46 &  FOR: 51.31 & Acc: 50.95 \\
        \bottomrule
    \end{tabular}
\end{table}

\begin{table}[t]
    \caption{Confusion matrix in the second \textit{Turing Test}. The volunteers are asked to estimate whether the generated style successfully imitates the target style or not. The values include Recall (Rec), Precision (Pre), False Positive Rate (FPR), False Omission Rate (FOR) and Accuracy (Acc). Although it is a more strict test than the first one, the result is reasonable and acceptable.}
    \label{table:human-2}
    \centering
    \begin{tabular}{c c c c}
    \toprule
        \multirow{2}{*}{Actual} & \multicolumn{2}{c}{Predicted}\\
        \cmidrule{2-3}
         & Real & Fake&\\
         \midrule
         Genuine & 35.95  & 14.05 & Rec: 71.90 \\
         Generated & 30.70 & 19.30 &  FPR: 61.40 \\
        &  Pre: 53.94 &  FOR: 42.13 & Acc: 55.25 \\
        \bottomrule
    \end{tabular}
\end{table}

Furthermore, we report another result for reference by raising the quality requirements of the generated images. The generated image is required not only to be like a real handwriting image, but also to successfully imitate a handwriting style. Specifically, we gave the volunteers three real handwriting images in the same style and asked them to determine whether the style of another image is the same or not. Every time the image used for classification was randomly collected from the existing images and the generated images in the same style. We collected 50 valid responses from every volunteer, namely 2,000 responses in total. As shown in the Table~\ref{table:human-2}, although the accuracy increases from 50.95\% (in Table~\ref{table:human-1}) to 55.25\%, we surprisingly find that it is still close to a random binary classification. Note that the setting (additional handwriting style requirements) is much stricter than the first human experiment. This experiment again verifies that the samples generated by our method can significantly imitate the real handwriting of human beings.

\section{Conclusion}

In this paper, we have presented SLOGAN for enriching the handwriting training samples for robust recognition. We synthesize handwriting data from the perspective of parameterizing style and controlling the parameters to generate new styles. This is achieved by using a style bank to parameterize a handwritten style as a latent vector, which is taken by the generator as a style prior to imitate the specific handwriting style. The joint training of the style bank and the generator requires only the supervision of the writer ID. By this way, it is possible to generate diverse handwriting styles by simply manipulating the latent vector. 

Another highlight of our method is that we propose dual discriminators specifically designed for text string images to provide relatively comprehensive supervision, thereby enabling text string image synthesis in an image-to-image manner. There are benefits of transferring a printed style image to a handwriting style image. First, it can generate a word or sentence out of the vocabulary by simply changing the input text content. Second, different spatial arrangements of the text string on the input image result in consistent effects on the output image, which indicates that our generator can synthesize text images of arbitrary shapes. 

Extensive experiments reveal the superiority of our method in terms of the generated quality and diversity, and its contribution to the robust training of recognizers. Our data synthesis method can also complement data augmentation methods and further boost recognition performance. It is also notable that our method is potentially capable of domain-adaptation tasks. Finally, the \textit{Turing Test} shows that our artificially synthesized samples are quite plausible that they can cheat the human judgment. In the future, we will explore online hard example generation by studying the interpretability of the GAN latent space. By this way, more effective and specific handwriting samples for training may be available.

\section*{Acknowledgment}

This research was supported in part by NSFC (Grant No. 61936003) and GD-NSF (No. 2017A030312006).

\bibliographystyle{IEEEtran}
\bibliography{reference}

\newpage
\appendix
We present several sayings written via our method. Zoom in for better view.

\begin{figure}[h] 
\includegraphics[width=0.75\columnwidth,height=0.22\columnwidth]{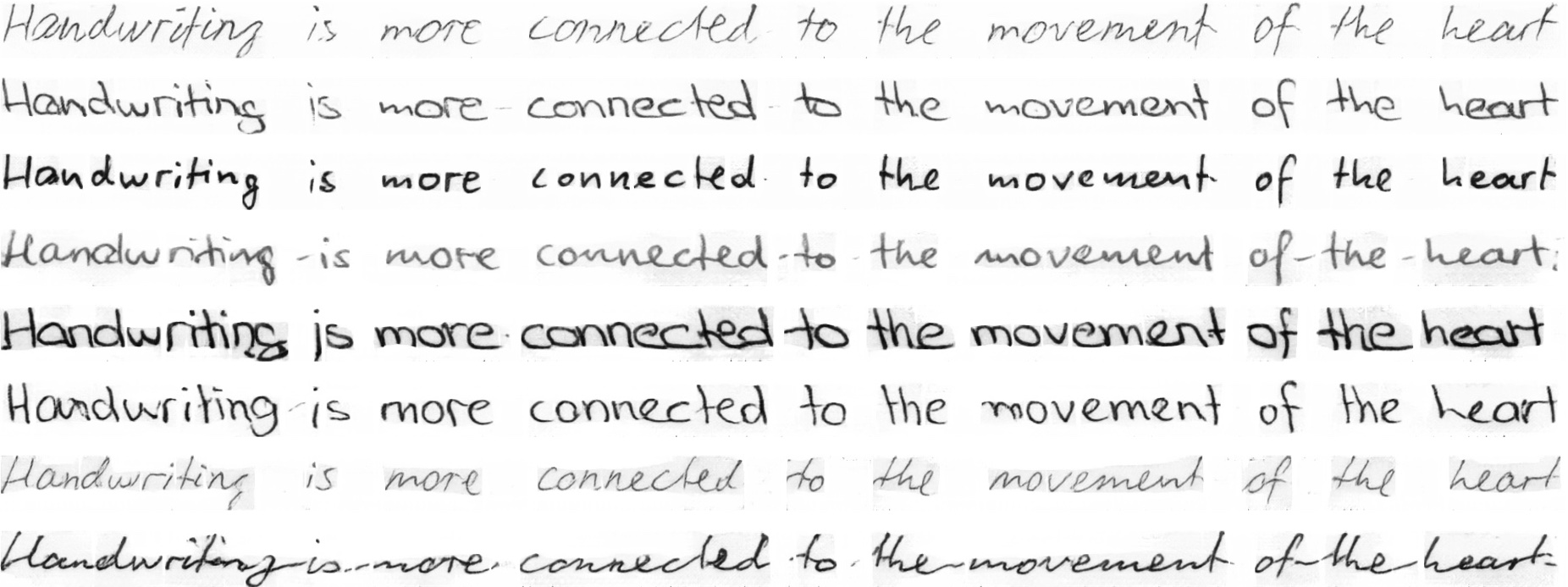} 
\caption{Handwriting is more connected to the movement of the heart.}
\end{figure}

\begin{figure}[h] 
\includegraphics[width=1.\columnwidth,height=0.22\columnwidth]{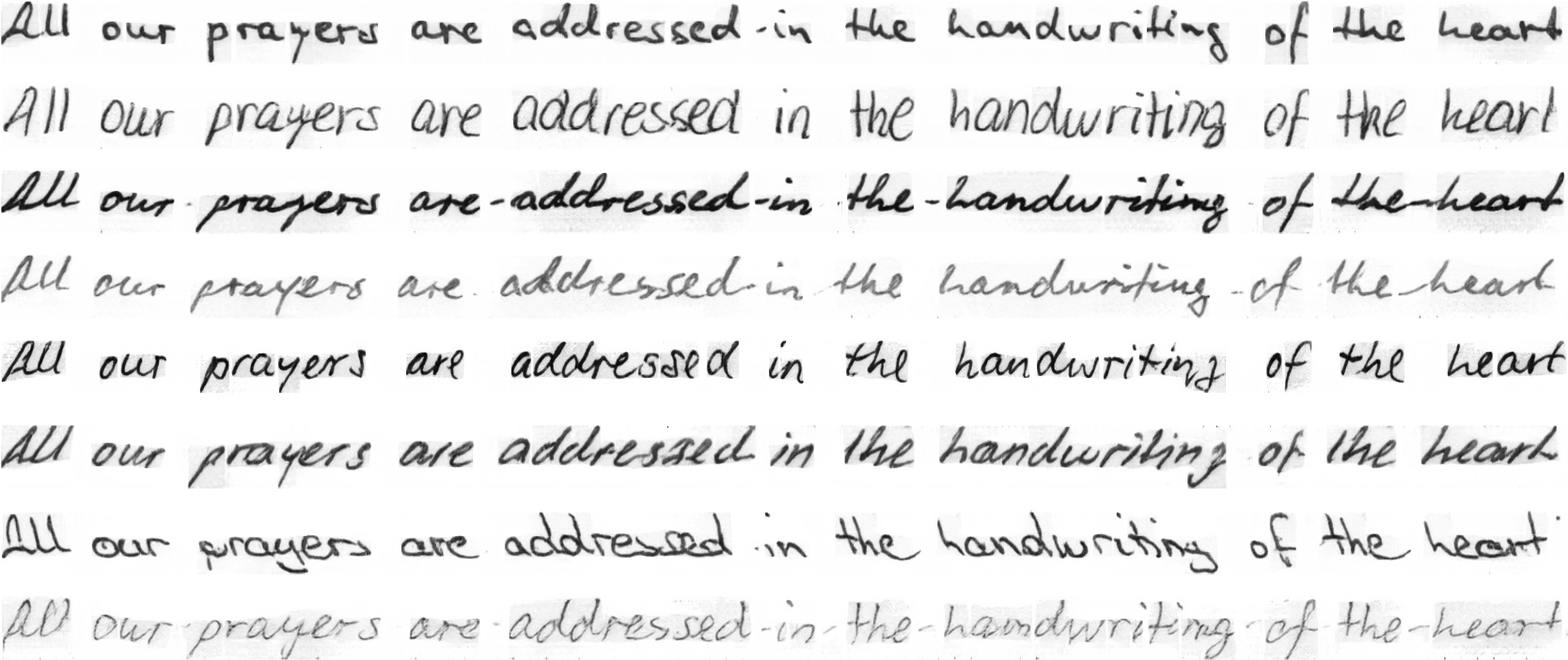} 
\caption{All our prayers are addressed in the handwriting of the heart.}
\end{figure}

\begin{figure}[h] 
\includegraphics[width=1.\columnwidth,height=0.22\columnwidth]{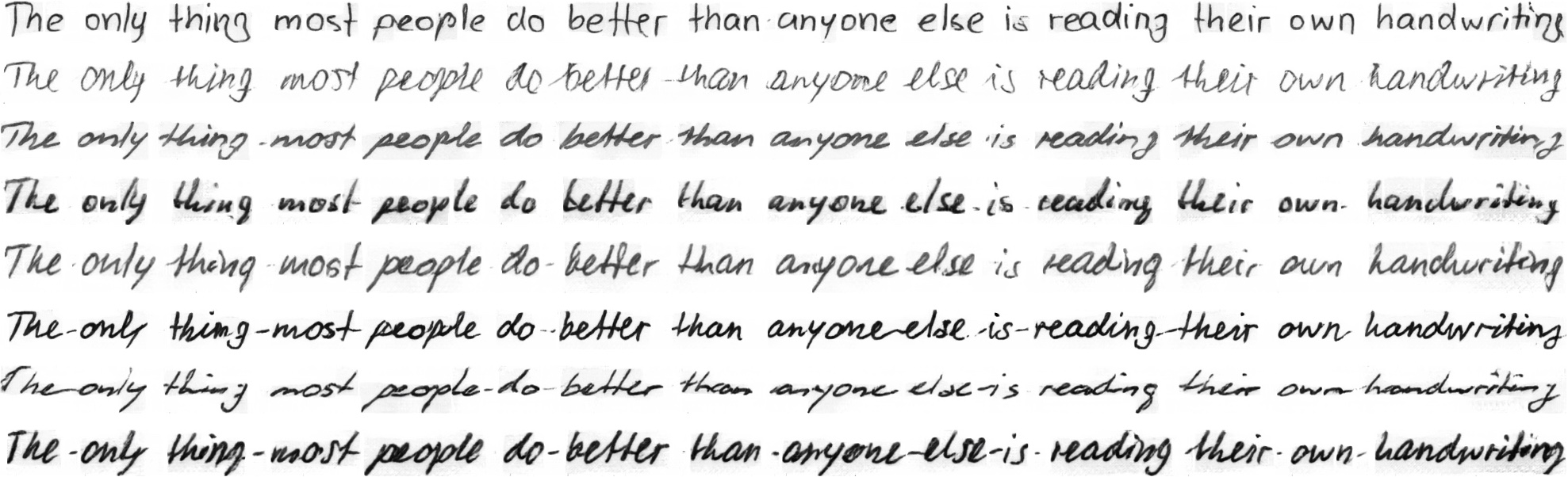} 
\caption{The only thing most people do better than anyone else is reading their own handwriting.}
\end{figure}

\begin{figure}[h] 
\includegraphics[width=1.\columnwidth,height=0.22\columnwidth]{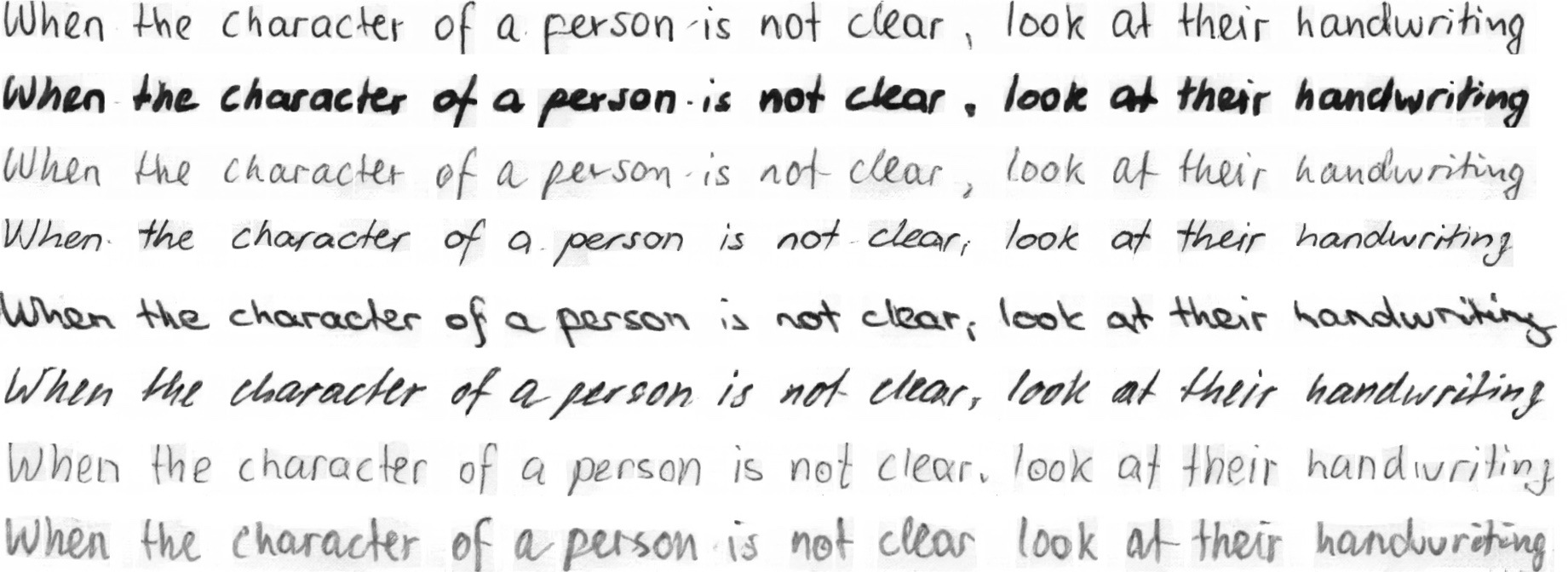} 
\caption{When the character of a person is not clear, look at their handwriting.}
\end{figure}




%
%
%

\end{document}